\PassOptionsToPackage{table,xcdraw}{xcolor}
\documentclass[11pt, a4paper, copyright, google]{google}

\usepackage[authoryear, sort&compress, round]{natbib}
\usepackage{xspace}
\usepackage{tikz}
\usepackage{tcolorbox}
\usepackage{fontawesome5}
\usetikzlibrary{shapes,arrows.meta,positioning,fit,backgrounds,calc}

\definecolor{brandblue}{RGB}{33, 78, 138}
\definecolor{brandteal}{RGB}{30, 130, 118}
\definecolor{subbox}{RGB}{238, 243, 248}

\definecolor{fiderr}{HTML}{F8D7DA}

\definecolor{qhigh}{HTML}{B7E4C7}
\definecolor{qmid}{HTML}{E8F6EC}
\newcommand{\qcell}[2]{\cellcolor{#1}#2}
\DeclareRobustCommand{\errbar}{%
  \tikz[baseline=-0.4ex]{%
    \node[anchor=east,inner sep=1.5pt,font=\scriptsize] at (0,0.07) {$0.0$};
    \shade[left color=qhigh, middle color=qmid, right color=fiderr] (0,-0.04) rectangle (1.4,0.18);
    \draw[black!30, line width=0.2pt] (0,-0.04) rectangle (1.4,0.18);
    \node[anchor=west,inner sep=1.5pt,font=\scriptsize] at (1.4,0.07) {$1.0$};
  }%
}
\DeclareRobustCommand{\realismbar}{%
  \tikz[baseline=-0.4ex]{%
    \node[anchor=east,inner sep=1.5pt,font=\scriptsize] at (0,0.07) {close to real};
    \shade[left color=qhigh, middle color=qmid, right color=fiderr] (0,-0.04) rectangle (1.4,0.18);
    \draw[black!30, line width=0.2pt] (0,-0.04) rectangle (1.4,0.18);
    \node[anchor=west,inner sep=1.5pt,font=\scriptsize] at (1.4,0.07) {departs from real};
  }%
}

\newcommand{\system}{{{\textsc{Synth\-Corp}}}\xspace}
\newcommand{\sps}{{{\textsc{SPs}}}\xspace}
\DeclareRobustCommand{\llmicon}{%
  \tikz[baseline=-0.3ex, scale=0.11, every node/.style={inner sep=0pt,outer sep=0pt}]{%
    \fill[brandblue] (0,1.25) .. controls (0.18,0.35) and (0.35,0.18) .. (1.25,0)
                     .. controls (0.35,-0.18) and (0.18,-0.35) .. (0,-1.25)
                     .. controls (-0.18,-0.35) and (-0.35,-0.18) .. (-1.25,0)
                     .. controls (-0.35,0.18) and (-0.18,0.35) .. cycle;
    \fill[brandteal] (1.15,1.15) ++(0,0.58) .. controls +(0.08,-0.42) and +(-0.42,0.08) .. +(0.58,-0.58)
                     .. controls +(-0.42,-0.08) and +(0.08,0.42) .. +(-0.58,-0.58)
                     .. controls +(-0.08,0.42) and +(0.42,-0.08) .. +(-0.58,0.58)
                     .. controls +(0.42,0.08) and +(-0.08,-0.42) .. cycle;
  }%
}

\newcommand*\circled[1]{\tikz[baseline=(char.base)]{
            \node[shape=circle,draw,inner sep=0.8pt] (char) {#1};}}

\renewcommand{\today}{Nov 2026}
\uselogo{}
\title{SynthCorp: A Platform for Creating Benchmarks for Data Agents}

\correspondingauthor{\{\{jianizhang, halevy\}@google.com}

\author[1]{Jiani Zhang}
\author[1]{Manav Garg}
\author[1]{Eugenie Lai}
\author[1]{Cheng Zhen}
\author[1]{Eva Sharma}
\author[1]{Yang Pan}
\author[1]{Yaming Lin}
\author[1]{Alon Halevy}

\affil[1]{Google}

\begin{abstract}
Current benchmarks for LLM-driven data agents saturate quickly and cannot target specific agent capabilities.
We present \system, a platform that creates customizable data estates and benchmarks by simulating how business events produce and modify corporate data over time.
Each generated data estate combines relational databases, a document store, and a data catalog, and pairs them with evaluation tasks and executable gold programs.
First, given a business scenario and target complexity dimensions, \system synthesizes reusable event types and compiles each into a program over predefined simulation primitives.
Second, \system schedules event instances on a virtual calendar, interleaving routine and disruption events so that tables, documents, and catalog entries remain consistent.
Third, \system executes the calendar to materialize the data estate and records every operation in a simulation log.
Fourth, \system generates analytical tasks and derives verified gold programs from the simulation log.

We use \system to generate a first version of evaluation benchmarks, containing seven data estates (up to 1K operational tables) and 256 tasks across five domains.
The generated data estates pass deterministic pipeline checks (such as foreign-key violations, undeclared writes, calendar recurrence violations, and database execution failures) with a near-zero error rate, introduce cross-asset table and document complexities, and cover all requested dimensions.
Evaluating three models across four data-context settings shows that the benchmarks separate both data-context settings and model tiers.
Auditing candidate queries against causal provenance in the simulation log verifies gold programs without manual annotation, including on hard multi-event composite tasks that expose reasoning failures in state-of-the-art agents.
We hope the community uses \system to generate more challenging and useful data estates and benchmarks that foster the development of enterprise data agents.
\end{abstract}

\begin{document}

\maketitle

\section{Introduction}
\label{sec:intro}

Autonomous data agents powered by LLMs are changing how enterprises handle data engineering and analytics~\citep{fu2025autonomous,zhu2025survey, jin2025elt,halevy2026toaster}. Building and improving these agents requires benchmarks that reflect the complexities of enterprise data estates, where thousands or millions of tables coexist with policy memos, business plans, and incomplete metadata catalogs. Researchers and agent builders, however, rarely have access to proprietary corporate data, which makes benchmark creation a primary bottleneck in agent development.

Existing data-agent benchmarks have three limitations~\citep{egg2025dabstep,lai2026kramabench,lei2026dacomp,ma2026can}. First, they provide fixed databases and static task sets that models saturate quickly. Second, developers cannot customize them to test specific agent capabilities, such as selecting authoritative tables from a warehouse with redundant copies, filtering out low-quality records, or resolving domain-specific business terms. Third, they start from a static data snapshot that does not record how or why the data reached that state. In a real enterprise, the difficulty of an analytical question often stems from that history, such as which table replaced a deprecated table in a migration or when a new accounting policy took effect. Annotators who only see the final snapshot have to invent questions and reverse-engineer reference answers, which is slow and error-prone when they deal with complex schemas and business rules~\citep{jin2026pervasive}.

\input{sections/fig_intro_example}

We present \system, a platform for creating customizable data estates and benchmarks for data agents. \system flips the relationship between agent building and benchmarking. Developers specify a business scenario and the data complexities they want to test, such as schema drift, and \system generates both the corporate data estate and the evaluation benchmark.
The \emph{data estate} consists of relational databases, a document store (e.g., business plans, policy memos, and meeting notes), and an often incomplete data catalog.
The benchmark consists of evaluation tasks and the corresponding executable \emph{gold programs}. Figure~\ref{fig:intro_example} illustrates this workflow on a specialty coffee shop with the requested schema drift and implicit business rules complexities. From these inputs, \system produces versioned tables, a series of documents including menu memos and staff policy, and several evaluation tasks and their verified gold programs.

Generating such a data estate is challenging because real-world data complexities arise from \emph{business events} that unfold over time and affect several data assets at once. Consider a schema migration in a coffee shop that starts recording milk choices on Day~3. This single event may leave behind a pre-migration snapshot frozen at the cutover date, an authoritative \texttt{transactions} table whose new \texttt{milk\_choice} column is \texttt{NULL} for rows before Day~3, a policy memo announcing the menu upgrade, and downstream views that query the new column only after the cutover. Because all of these changes come from the same event, the tables, documents, and views agree on what changed and when. Approaches that directly generate a final database snapshot~\citep{li2025omnisql, habibollah2026synsql,huang2025crmarenapro,xu2025theagentcompany,dai2025lakegen} have to coordinate all of these cross-asset constraints at once, which breaks down as the schema grows. Approaches that inject synthetic noise into a clean database~\citep{abdelaal2023rein,chang2023dr,luoma2025snails,zhang2026agentic} corrupt tables or documents in isolation and leave no business context that explains them.

The key idea of \system is to build a data estate by simulating the business events that create and modify corporate data over time. \system models two kinds of events and schedules them on a shared virtual calendar. \emph{Routine events} describe recurring operations, such as daily customer orders or weekly inventory audits, and populate the operational tables and reports. \emph{Disruption events} introduce structural or policy changes, such as upgrading a table schema, freezing an audit snapshot, or issuing a new revenue policy. Executing both kinds of events in an order that mirrors the life of the enterprise keeps tables, documents, and catalog entries mutually consistent. Data complexities emerge both from planned disruptions and from routine operations, such as dated audit snapshots that accumulate alongside authoritative tables in the warehouse.

Making such a simulation reliable by leveraging LLMs rests on three design principles. First, users cannot manually specify thousands of low-level event instances or anticipate every operation needed to produce a data complexity. Users provide a concise business scenario and select target complexity dimensions and then \system synthesizes reusable routine and disruption \emph{event types} in natural language for human inspection and refinement. Separating event types from the simulation schedule lets \system define and review each event type once and instantiate it as many times as necessary across the virtual calendar.
Second, \system compiles each high-level event type into a program over predefined \emph{simulation primitives} (\sps). The simulation primitives are the operations that create and modify the simulated data estate. Each primitive performs a single bounded operation that an LLM executes reliably, while explicit input and output operands let the compiler and scheduler verify cross-event consistency before execution. In a sense, these primitives establish a \emph{boundary of trust} delineating where we feel that we can delegate the rest of the work to the LLM reliably.
Third, \system records the complete provenance of operations on the data estate in a \emph{simulation log}. \system hides this log from evaluated agents and uses it during benchmark creation to determine when a synthesized question is answerable from the observable data estate and to verify its reference gold program. During execution, \system binds \emph{benchmark hints} to created tables and documents, augments them with post-hoc detectors over the log, and composes verified atomic tasks into multi-complexity tasks. An LLM agent then derives each gold program step by step against the simulation log, documents, and database.

We validate four advantages of \system by generating seven data estates and 256 tasks in five domains (retail at 10, 100, and 1K operational tables, music, banking, logistics, and insurance) across three development scenarios and four held-out scenarios.
(1)~\emph{Reliable and regenerable}. \system generates a new data estate and benchmark in minutes to hours with a near-zero error rate on deterministic pipeline correctness checks, so developers can create fresh benchmarks as models saturate existing ones. The six smaller data estates pass every deterministic check on their compiled programs, calendars, and databases, and Retail-1000 passes all static program and calendar checks with a 0.2\% runtime query failure rate when recurring refresh queries read from empty upstream tables.
(2)~\emph{Targeted and cross-asset}. Developers choose the business domain, warehouse scale, and complexity dimensions to test, and every generated benchmark covers all requested dimensions. As business events update tables, documents, and catalog entries together, authoritative tables coexist with backups, staging tables, and reporting snapshots on up to 68\% of tasks, 64\% face superseded statements across documents and the glossary, and 80\% combine SQL queries with document or catalog evidence.
(3)~\emph{Discriminative}. Evaluating two open-weight models and a frontier model across four data-context settings shows that the generated benchmarks separate both data-context settings and model tiers. Because \system embeds business rules in documents and the glossary, schema-only prompting fails on these tasks, whereas retrieving a single document raises overall accuracy by 9 to 17 percentage points and multi-step agentic search adds another 12 to 20 points.
(4)~\emph{Verifiable and challenging}. Every complexity traces to a logged business event, so \system derives gold programs without manual annotation, and state-of-the-art agents match them on up to 94\% of runs after accounting for format differences. On harder multi-event composite tasks, causal provenance in the simulation log refines ambiguous questions and corrects candidate gold programs, producing verified tasks that expose multi-hop and point-in-time reasoning failures across all evaluated models.

\section{Related Work}
\label{sec:related_work}

\providecommand{\yes}{\ding{51}}
\providecommand{\no}{\ding{55}}
\providecommand{\half}{$\circ$}
\providecommand{\na}{--}
\newcommand{\hdr}[1]{\begin{tabular}[b]{@{}c@{}}#1\end{tabular}}
\newcommand{\lwork}[2]{#1\\[-1pt]{\footnotesize #2}}

\begin{table}[t]
\centering\small
\caption{Comparison of \system with related work. Assets are tables (T), documents (D), and catalog (C). \yes~full, \half~partial, \no~none, and \na~logs only.}
\label{tab:related_work}
\setlength{\tabcolsep}{0pt}
\renewcommand{\arraystretch}{1.18}
\begin{tabular*}{\textwidth}{@{\extracolsep{\fill}}>{\raggedright\arraybackslash}p{6.65cm}cccccc@{}}
\toprule
\textbf{Line of work} & \hdr{\textbf{Controllable}\\\textbf{Complexity}} & \hdr{\textbf{Schema}\\\textbf{Evolution}} & \hdr{\textbf{Virtual}\\\textbf{Calendar}} & \textbf{Assets} & \hdr{\textbf{Data}\\\textbf{Provenance}} & \hdr{\textbf{Verified}\\\textbf{Tasks}} \\
\midrule
\lwork{Curated benchmarks}{\citep{lai2026kramabench,lei2025spider,chen2024beaver}} & \no & \no & \no & T, D & \no & \no \\
\addlinespace[1.5pt]
\lwork{Enterprise simulators}{\citep{flynt2026orgforge,zeng2026minicorp}} & \no & \no & \yes & T, D & \yes & \no \\
\addlinespace[1.5pt]
\lwork{Enterprise sandboxes}{\citep{huang2025crmarenapro,xu2025theagentcompany,gruenbaum2026era}} & \half & \no & \half & T, D & \half & \half \\
\addlinespace[1.5pt]
\lwork{Process simulation}{\citep{burattin2016plg2,camargo2020simod,kirchdorfer2024agentsimulator}} & \no & \no & \yes & \na & \half & \no \\
\addlinespace[1.5pt]
\lwork{Database generators}{\citep{sanghi2018hydra,patki2016sdv,pang2024clavaddpm,yang2024sense,habibollah2026synsql,dai2025lakegen}} & \half & \no & \no & T & \no & \half \\
\addlinespace[1.5pt]
\lwork{Corruption \& drift}{\citep{arocena2015bart,zhang2026agentic,zhang2025evoschema}} & \yes & \half & \no & T & \half & \half \\
\midrule
\textbf{\system (this work)} & \yes & \yes & \yes & T, D, C & \yes & \yes \\
\bottomrule
\end{tabular*}
\end{table}

\textbf{Data Agent Benchmarks.}
Recent benchmarks test data agents on multi-database querying, exploratory question answering, and text-to-SQL workflows~\citep{lai2026kramabench,ma2026can,lei2025spider,wang2026lakeqa,zhang2026coda,lao2025sembench,jing2025dsbench,jin2025elt,chen2024beaver}. Table~\ref{tab:related_work} compares \system with representative benchmarks and generators.
Most of these benchmarks collect static public databases or repositories and rely on human annotation to author questions and gold programs~\citep{northcutt2021pervasive,weber2024varierr}. Because static snapshots lack a record of how the data evolved, developers cannot adjust the difficulty of these benchmarks or target specific agent capabilities.

\noindent \textbf{Synthetic Data and Corruption.}
Prior database generators produce test data for query engines~\citep{gray1994quickly,touchstone18,sanghi2018hydra}, match the statistical distributions of existing tables~\citep{xu2019modeling,zhang2024mixed,patki2016sdv,pang2024clavaddpm}, or synthesize text-to-SQL training pairs and benchmark instances~\citep{long2024llms,li2025omnisql,yang2024sense,habibollah2026synsql,dai2025lakegen}.
All of these systems generate a single snapshot of a fixed schema. A complementary line of work injects cell errors~\citep{arocena2015bart,zhang2026agentic} or perturbs database schemas~\citep{zhang2025evoschema}. However, post-hoc corruption modifies individual tables in isolation and omits the policy memos, views, and historical snapshots that real business events leave behind.

\noindent \textbf{Enterprise and Process Simulation.}
Recent simulators model organizations and multi-agent interactions~\citep{park2023generative,ni2026survey}. Enterprise simulators record a deterministic event log from which an LLM renders documents~\citep{flynt2026orgforge}, or run role-based agent teams against a simulated market~\citep{zeng2026minicorp}. These systems use hand-coded domain simulators, keep database schemas fixed, and generate text corpora without verified analytical tasks. Enterprise sandboxes~\citep{huang2025crmarenapro,xu2025theagentcompany,gruenbaum2026era} populate fixed schemas and evaluate agents with hand-written task templates or inject defects from a static manifest. Business-process simulators~\citep{burattin2016plg2,camargo2020simod,kirchdorfer2024agentsimulator} emit event logs for process mining and do not materialize relational tables, documents, or metadata catalogs.

\noindent \textbf{LLMs as World Models.}
Prior work uses pretrained LLMs as world models to predict state transitions or synthesize executable simulators~\citep{hao2023rap,tang2024worldcoder,kambhampati2024llmmodulo}. \system builds on this direction by using an LLM as an enterprise world model that expands a short scenario into business event types, workflows, and distributions, and compiles them into bounded simulation primitives.

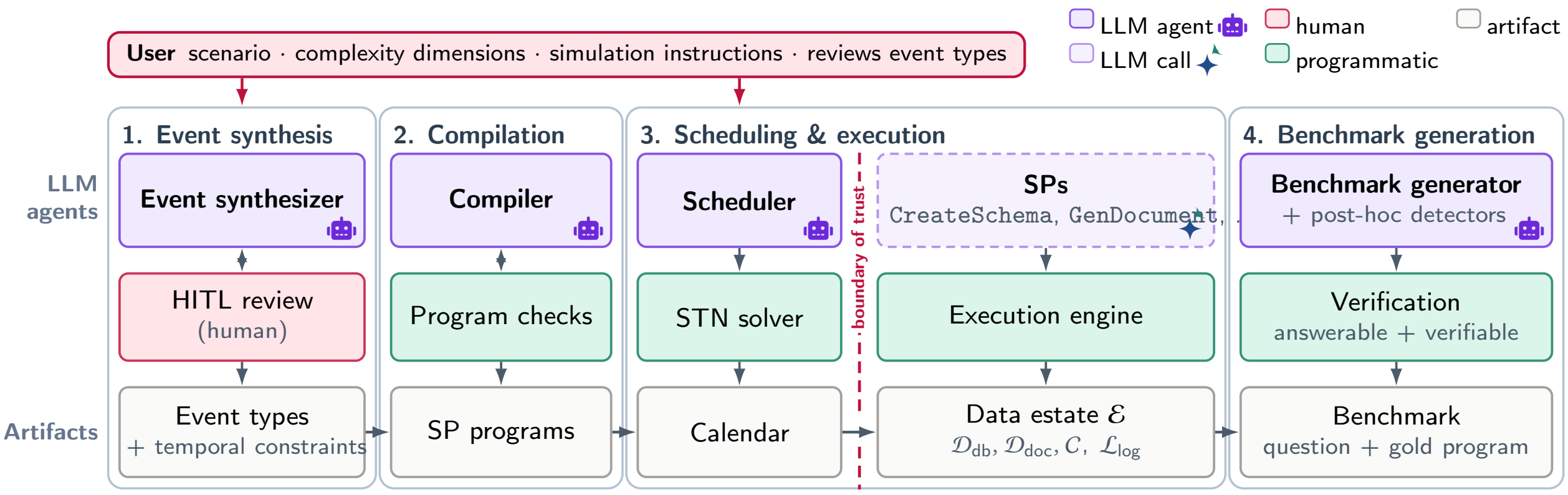
\begin{figure}[t]
 \centering
 \providecolor{flane}{HTML}{64748B}%
 \providecolor{fframe}{HTML}{B4C3D2}%
 \providecolor{ftitle}{HTML}{2C3E50}%
 \providecolor{fllm}{HTML}{6D28D9}\providecolor{fllmbg}{HTML}{EFEAFE}
 \providecolor{fdet}{HTML}{047857}\providecolor{fdetbg}{HTML}{DCF5EA}
 \providecolor{fhum}{HTML}{BE123C}\providecolor{fhumbg}{HTML}{FFE4E8}
 \providecolor{fart}{HTML}{57534E}\providecolor{fartbg}{HTML}{FAFAF9}
 \ifdefined\colgap\else
 \newlength{\colgap}\newlength{\rowA}\newlength{\rowB}\newlength{\rowC}\newlength{\rowgap}%
 \fi
 \setlength{\colgap}{0.32cm}%
 \setlength{\rowA}{1.12cm}%
 \setlength{\rowB}{1.05cm}%
 \setlength{\rowC}{1.08cm}%
 \setlength{\rowgap}{0.32cm}%
\providecommand{\sub}{}\renewcommand{\sub}{\scriptsize\color{ftitle!85}}
\providecommand{\figllmicon}{}\renewcommand{\figllmicon}{{\color{fllm}\faRobot}}
\providecommand{\figgenicon}{}\renewcommand{\figgenicon}{{\color{fllm}\llmicon}}
\resizebox{\textwidth}{!}{%
\begin{tikzpicture}[
 font=\sffamily\footnotesize,
 cell/.style={rectangle, rounded corners=3pt, line width=0.8pt, inner xsep=4pt, inner ysep=3pt,
 align=center, anchor=north west},
 llmcell/.style={cell, draw=fllm!80, fill=fllmbg, minimum height=\rowA},
 llmlite/.style={cell, draw=fllm!50, fill=fllmbg!60, minimum height=\rowA, densely dashed},
 detcell/.style={cell, draw=fdet!80, fill=fdetbg, minimum height=\rowB},
 humcell/.style={cell, draw=fhum!85, fill=fhumbg, minimum height=\rowB},
 artcell/.style={cell, draw=fart!55, fill=fartbg, minimum height=\rowC},
 stagetitle/.style={font=\sffamily\footnotesize\bfseries, text=ftitle, anchor=north west, inner sep=0pt},
 stageframe/.style={rectangle, rounded corners=5pt, draw=fframe, line width=0.75pt,
 inner xsep=3.5pt, inner ysep=3.5pt},
 lanelabel/.style={font=\sffamily\scriptsize\bfseries, text=flane, anchor=east,
 inner sep=1pt, align=right},
 band/.style={rectangle, rounded corners=4pt, line width=0.9pt, inner xsep=6pt, inner ysep=3.5pt,
 align=flush left, font=\sffamily\footnotesize},
 userband/.style={band, draw=fhum, fill=fhumbg, anchor=south west},
 arrow/.style={-{Latex[length=2.1mm,width=1.45mm]}, line width=0.95pt, draw=ftitle!85},
 varrow/.style={arrow, line width=0.8pt},
 loop/.style={{Latex[length=1.7mm,width=1.25mm]}-{Latex[length=1.7mm,width=1.25mm]}, line width=0.8pt, draw=ftitle!85},
 uarrow/.style={arrow, draw=fhum},
 swatch/.style={rectangle, rounded corners=1.5pt, minimum width=0.28cm, minimum height=0.22cm, inner sep=0pt, line width=0.6pt},
]

\def\wA{2.95cm}\def\wB{2.65cm}\def\wC{2.45cm}\def\wD{4.05cm}\def\wE{3.75cm}
\def\titleh{0.42cm}
\coordinate (yA) at (0,0);
\coordinate (yB) at ($(yA)+(0,-\rowA-\rowgap)$);
\coordinate (yC) at ($(yB)+(0,-\rowB-\rowgap)$);
\coordinate (xA) at (1.35cm,0);
\coordinate (xB) at ($(xA)+(\wA,0)+(\colgap,0)$);
\coordinate (xC) at ($(xB)+(\wB,0)+(\colgap,0)$);
\coordinate (xD) at ($(xC)+(\wC,0)+(0.44cm,0)$);
\coordinate (xE) at ($(xD)+(\wD,0)+(\colgap,0)$);

\node[lanelabel] at ($(xA)+(-0.20cm,0)+(0,-0.5\rowA)$) {LLM\\agents};
\node[lanelabel] at ($(xA)+(-0.20cm,0)+(0,-\rowA-\rowgap-0.5\rowB)$) {};
\node[lanelabel] at ($(xA)+(-0.20cm,0)+(0,-\rowA-\rowB-2\rowgap-0.5\rowC)$) {Artifacts};

\node[llmcell, text width=\wA-8pt] (a1) at (xA |- yA) {\textbf{Event synthesizer}};
\node[humcell, text width=\wA-8pt] (b1) at (xA |- yB) {HITL review\\[-1pt]{\sub (human)}};
\node[artcell, text width=\wA-8pt] (c1) at (xA |- yC) {Event types\\[-1pt]{\sub \mbox{+ temporal constraints}}};

\node[llmcell, text width=\wB-8pt] (a2) at (xB |- yA) {\textbf{Compiler}};
\node[detcell, text width=\wB-8pt] (b2) at (xB |- yB) {Program checks};
\node[artcell, text width=\wB-8pt] (c2) at (xB |- yC) {SP programs};

\node[llmcell, text width=\wC-8pt] (a3) at (xC |- yA) {\textbf{Scheduler}};
\node[detcell, text width=\wC-8pt] (b3) at (xC |- yB) {STN solver};
\node[artcell, text width=\wC-8pt] (c3) at (xC |- yC) {Calendar};

\node[llmlite, text width=\wD-8pt] (a4) at (xD |- yA) {\textbf{\sps}\\[-1pt]{\sub \mbox{\texttt{CreateSchema}, \texttt{GenDocument}, \ldots}}};
\node[detcell, text width=\wD-8pt] (b4) at (xD |- yB) {Execution engine};
\node[artcell, text width=\wD-8pt] (c4) at (xD |- yC) {Data estate $\mathcal{E}$\\[-1pt]{\sub $\mathcal{D}_{\text{db}}, \mathcal{D}_{\text{doc}}, \mathcal{C}$, $\mathcal{L}_{\text{log}}$}};

\node[llmcell, text width=\wE-8pt] (a5) at (xE |- yA) {\textbf{Benchmark generator}\\[-1pt]{\sub + post-hoc detectors}};
\node[detcell, text width=\wE-8pt] (b5) at (xE |- yB) {Verification\\[-1pt]{\sub answerable + verifiable}};
\node[artcell, text width=\wE-8pt] (c5) at (xE |- yC) {Benchmark\\[-1pt]{\sub question + gold program}};

\foreach \i in {1,2,3,5}{
 \node[anchor=south east, inner sep=0pt, font=\scriptsize] at ($(a\i.south east)+(-3pt,2.5pt)$) {\figllmicon};
}
\node[anchor=south east, inner sep=0pt, font=\scriptsize] at ($(a4.south east)+(-3pt,2.5pt)$) {\figgenicon};

\coordinate (t1) at ($(a1.north west)+(0,\titleh)$);
\coordinate (t2) at ($(a2.north west)+(0,\titleh)$);
\coordinate (t3) at ($(a3.north west)+(0,\titleh)$);
\coordinate (t5) at ($(a5.north west)+(0,\titleh)$);
\begin{scope}[on background layer]
 \node[stageframe, fit=(a1)(c1)(t1)] (f1) {};
 \node[stageframe, fit=(a2)(c2)(t2)] (f2) {};
 \node[stageframe, fit=(a3)(c3)(a4)(c4)(t3)] (f3) {};
 \node[stageframe, fit=(a5)(c5)(t5)] (f5) {};
\end{scope}
\node[stagetitle] at ($(a1.north west)+(1pt,\titleh-2pt)$) {1. Event synthesis};
\node[stagetitle] at ($(a2.north west)+(1pt,\titleh-2pt)$) {2. Compilation};
\node[stagetitle] at ($(a3.north west)+(1pt,\titleh-2pt)$) {3. Scheduling \& execution};
\node[stagetitle] at ($(a5.north west)+(1pt,\titleh-2pt)$) {4. Benchmark generation};

\draw[loop] (a1.south) -- (b1.north);
\draw[loop] (a2.south) -- (b2.north);
\foreach \i in {3,4,5}{ \draw[varrow] (a\i.south) -- (b\i.north); }
\foreach \i in {1,2,3,4,5}{ \draw[varrow] (b\i.south) -- (c\i.north); }

\coordinate (bx) at ($(a3.east)!0.5!(a4.west)$);
\coordinate (cmid) at ($(c1.north)!0.5!(c1.south)$);
\draw[fhum, line width=1.0pt, dash pattern=on 4pt off 2.5pt]
 (bx |- a3.north) -- ($(bx |- cmid)+(0,5pt)$)
 ($(bx |- cmid)+(0,-5pt)$) -- (bx |- f3.south);
\coordinate (btrustY) at ($(a3.north)!0.5!(b3.south)$);
\node[font=\sffamily\fontsize{6.5}{8}\selectfont\bfseries, text=fhum, fill=white, inner sep=1.2pt, rotate=90, anchor=center]
 at (bx |- btrustY) {boundary of trust};

\foreach \i/\j in {1/2,2/3,3/4,4/5}{ \draw[arrow] (c\i.east |- cmid) -- (c\j.west |- cmid); }

\coordinate (topy) at ($(f1.north)+(0,0.34cm)$);
\node[userband, font=\sffamily\scriptsize, inner xsep=6pt, inner ysep=4pt] (user) at (f1.west |- topy) {%
 \textbf{User}\enspace scenario $\cdot$ complexity dimensions $\cdot$ simulation instructions $\cdot$ reviews event types};
\draw[uarrow] (user.south -| a1.north) -- (f1.north -| a1.north);
\draw[uarrow] (user.south -| a3.north) -- (f3.north -| a3.north);

\newcommand{\lsw}[2]{\tikz[baseline=-0.5ex]{\node[swatch, draw=#1, fill=#2] {};}}
\node[anchor=south east, inner sep=0pt, font=\sffamily\scriptsize] at ($(f5.east |- user.south)+(-1pt,0.0cm)$) {%
 \setlength{\tabcolsep}{2pt}\renewcommand{\arraystretch}{1.12}%
 \begin{tabular}{@{}l@{\hspace{2pt}}l@{\hspace{6pt}}l@{\hspace{2pt}}l@{\hspace{6pt}}l@{\hspace{2pt}}l@{}}
 \lsw{fllm!80}{fllmbg} & LLM agent\,\figllmicon & \lsw{fhum!85}{fhumbg} & human & \lsw{fart!55}{fartbg} & artifact\\
 \lsw{fllm!50}{fllmbg!60} & LLM call\,\figgenicon & \lsw{fdet!80}{fdetbg} & programmatic & &
 \end{tabular}};

\end{tikzpicture}%
}
 \caption{\textbf{The \system pipeline.} The event synthesizer (Stage~1), the compiler (Stage~2), the scheduler and the execution engine (Stage~3), and the benchmark generator (Stage~4) each pair LLM generation with human or programmatic verification and pass a validated artifact to the next stage.}
 \label{fig:synthcorp_pipeline}
 \vspace{-2mm}
\end{figure}

\section{The \system Design}
\label{sec:system}

Figure~\ref{fig:synthcorp_pipeline} shows the four stages of \system. First, \system turns a business scenario and requested complexity dimensions into event types for user review. Second, it compiles each approved event type into an SP program. Third, it places event instances on a virtual calendar from the user's simulation instructions and executes them to materialize the data estate and simulation log. Fourth, it derives the benchmark with diverse analytic tasks and gold programs from the data estate and log.

Figure~\ref{fig:intro_example} traces this pipeline on a 14-day specialty coffee shop simulation. From the scenario and two requested complexities, \system synthesizes a daily-sales routine event type and two disruption event types (a Day~3 menu upgrade and a Day~10 staff-drink policy), compiles them into SP programs, and schedules them across the 14-day calendar. Execution materializes a frozen \texttt{transactions\_v1} snapshot, an authoritative \texttt{transactions} table with unbackfilled \texttt{milk\_choice} values and \$0 staff drinks, and two policy memos, while logging provenance and binding benchmark hints to construct atomic and composite tasks. We now define each stage's inputs, outputs, and guarantees.

\subsection{The Simulated Data Estate}
\label{sec:data_estate}

We model a data estate as $\mathcal{E} = \langle \mathcal{D}_{\text{db}}, \mathcal{D}_{\text{doc}}, \mathcal{C}, \mathcal{L}_{\text{log}} \rangle$.

\smallskip
\noindent {\bf Relational databases ($\mathcal{D}_{\text{db}}$).} Operational (OLTP) databases and analytical (OLAP) warehouses. $\mathcal{D}_{\text{db}}$ keeps all created artifacts unless an event deletes them, including tables orphaned by a migration, added columns whose historical rows remain \texttt{NULL}, and ad-hoc materialized views and analyst views.

\smallskip
\noindent {\bf A document store ($\mathcal{D}_{\text{doc}}$).} Unstructured enterprise text, including business plans, policy memos, meeting notes, and supplier contracts. Critical rules (such as the rationale behind a schema change) often reside only here, requiring agents to consult $\mathcal{D}_{\text{doc}}$.

\smallskip
\noindent {\bf A data catalog ($\mathcal{C}$).} Metadata indexing tables in $\mathcal{D}_{\text{db}}$ and documents in $\mathcal{D}_{\text{doc}}$. $\mathcal{C}$ stores schemas, types, descriptions, optional lineage and update frequency, and a business glossary of domain terms, acronyms, and status codes that maps question terms to schema elements.
In real enterprises, metadata catalogs are rarely complete. Organizations frequently leave new tables undocumented or keep business definitions in documents that never reach the catalog. Data engineering teams refer to recovering and linking this missing metadata as \textit{catalog enhancement}. To test incomplete catalogs, \system can store business rules in $\mathcal{D}_{\text{doc}}$ and omit them from $\mathcal{C}$.

\smallskip
\noindent {\bf A simulation log ($\mathcal{L}_{\text{log}}$).} A record of every operation applied to the data estate and its rationale. \system hides $\mathcal{L}_{\text{log}}$ from evaluated agents and uses it only during benchmark generation to construct answerable tasks and verify gold programs.

\subsection{Event Types as the Abstraction Layer}\label{sec:event}

In Stage~1, given a business scenario and target data complexities, the event synthesizer produces a set of \emph{event types} that capture corporate operations at a compact and reusable level of abstraction. Routine events (such as serving customers, replenishing inventory, and running weekly audits) repeat frequently and populate operational tables and reports. Disruption events (such as upgrading a schema, revising an accounting policy, or migrating a pipeline) occur at scheduled points on the calendar and introduce data complexities.
Separating event types from instances lets \system compile each event type once and instantiate it across the horizon.

\begin{table}[t]
\centering\small
\caption{Complexity dimensions and representative events.}
\label{tab:complexity_dimensions}
\setlength{\tabcolsep}{6pt}
\renewcommand{\arraystretch}{1.15}
\begin{tabular}{@{}>{\raggedright\arraybackslash}p{3.9cm}>{\raggedright\arraybackslash}p{\dimexpr\textwidth-3.9cm-2\tabcolsep\relax}@{}}
\toprule
\textbf{Dimension} & \textbf{Effect on the data estate and example event} \\
\midrule
\textbf{Business glossary} & Domain terms or codes require definitions outside the schema, e.g., \texttt{PublishGlossaryPolicyEvent} defines POS status code 4 as a voided order. \\
\addlinespace[1.5pt]
\textbf{Implicit business rules} & Documents state rules that the schema does not enforce, e.g., \texttt{ReviseDiscountPolicy} writes a memo excluding staff-discounted sales from revenue. \\
\addlinespace[1.5pt]
\textbf{Multi-granularity} & Transaction rows coexist with coarser summaries, e.g., \texttt{EstablishDailyRollup} creates daily register summaries. \\
\addlinespace[1.5pt]
\textbf{Table redundancy} & Frozen snapshots and redundant copies remain beside authoritative tables, e.g., \texttt{AnalystAuditSnapshot} retains an audit snapshot. \\
\addlinespace[1.5pt]
\textbf{Schema drift} & Schema changes leave historical rows without new attributes, e.g., \texttt{UpgradeMenuEvent} freezes the snapshot and adds \texttt{milk\_\allowbreak choice} to the authoritative table. \\
\bottomrule
\end{tabular}
\end{table}

Each event type pairs a natural-language process description with four declarations.
First, \emph{parameters} (such as business hours and arrival rates in \texttt{RunCafeEvent}) specify constants, distributions, or references to parameters of other event types.
Second, \emph{recurrence} specifies one-time, periodic, or continuous execution across the horizon.
Third, \emph{temporal constraints} order instance intervals relative to other event types (e.g., \texttt{Setup\allowbreak Cafe\allowbreak Event} precedes \texttt{Run\allowbreak Cafe\allowbreak Event}).
Fourth, \emph{benchmark hints} state in business terms what an analyst or agent needs to know to query the affected data accurately, such as a maintained invariant, metric formula, or schema cutover.

\subsubsection{Temporal constraints}
\label{sec:temporal_constraints}

Temporal constraints specify when events may occur relative to one another. An instance $A$ occupies an interval $I_A=[s_A,f_A]$ with start and finish times $s_A$ and $f_A$ on the virtual calendar. \system adopts seven interval predicates based on Allen's interval algebra~\citep{allen1983maintaining}, each defined as a conjunction of endpoint inequalities that forms a simple temporal network:
(1)~\textbf{precedes}($A,B$), where $A$ finishes by the time $B$ starts ($f_A\leq s_B$),
(2)~\textbf{meets}($A,B$), where $A$ finishes when $B$ starts ($f_A=s_B$),
(3)~\textbf{within}($A,B$), where $A$ occurs entirely within $B$ ($s_B\leq s_A\land f_A\leq f_B$),
(4)~\textbf{overlaps}($A,B$), where $B$ starts during $A$ and finishes after $A$ ($s_A<s_B<f_A<f_B$),
(5)~\textbf{starts}($A,B$), where both start together and $A$ finishes by $B$ ($s_A=s_B\land f_A\leq f_B$),
(6)~\textbf{finishes}($A,B$), where both finish together and $A$ starts no earlier than $B$ ($f_A=f_B\land s_B\leq s_A$), and
(7)~\textbf{equals}($A,B$), where both share the same endpoints ($s_A=s_B\land f_A=f_B$).

\subsubsection{What the event set guarantees}
\label{sec:event_coherence}

Stage~1 ensures that the approved event types satisfy three properties before compilation.

\smallskip
\noindent\textbf{Domain fit.}
The routine event types model the operational workflows described in the user's scenario, and for each requested complexity dimension, a disruption event type introduces that complexity into the routine operations. Table~\ref{tab:complexity_dimensions} summarizes the complexity dimensions and representative events.

\smallskip
\noindent\textbf{Coherence.}
Even when individual event types look plausible on their own, a set of event types can fail in three ways. First, event types can overlap or contradict one another. For instance, two routine events might both record customer purchases and double-count revenue, or a setup event might define five menu items while the daily sales event samples from twelve. Second, a disruption event can become disconnected if it modifies a workflow or table that no routine event creates. Third, temporal constraints can form cycles or place a disruption event outside the active window of routine operations. For example, if \texttt{UpgradeMenuEvent} runs before Day~1 or after Day~14 of sales, all transaction rows share a single schema and schema drift never occurs.

\smallskip
\noindent\textbf{User approval.}
\label{sec:event_hitl}
Because a brief scenario description leaves fine-grained business choices open (such as how the point-of-sale system marks voided orders), Stage~1 surfaces these choices as clarification questions and lets the user add, remove, or edit event types in natural language before compilation. We call this step \emph{human-in-the-loop (HITL) review}. Once approved, the event set serves as the specification against which we evaluate all downstream stages.

\begin{table}[t]
\centering\small
\caption{The twelve simulation primitives in $\mathcal{S}$. \llmicon\ marks the category whose primitives invoke an LLM.}
\label{tab:simulation_primitives}
\setlength{\tabcolsep}{5pt}
\renewcommand{\arraystretch}{1.12}
\begin{tabular}{@{\hspace*{0.4em}}>{\raggedright\arraybackslash}p{3.6cm}>{\raggedright\arraybackslash}p{1.6cm}>{\raggedright\arraybackslash}p{\dimexpr\textwidth-5.2cm-0.4em-4\tabcolsep\relax}@{}}
\toprule
\multicolumn{1}{@{}l}{\textbf{Primitive}} & \textbf{Data} & \textbf{Operation} \\
\specialrule{\lightrulewidth}{1.5pt}{0pt}
\rowcolor{subbox}\multicolumn{3}{@{}l}{\rule{0pt}{2.2ex}\textbf{Generative}\,\llmicon} \\
\addlinespace[1.5pt]
\texttt{CreateSchema} & $\mathcal{D}_{\text{db}}, \mathcal{C}$ &
Generates a table schema from a business description and registers it in $\mathcal{C}$. \\
\addlinespace[1.5pt]
\texttt{GenDocument} & $\mathcal{D}_{\text{doc}}, \mathcal{C}$ &
Generates a document, inserts it into $\mathcal{D}_{\text{doc}}$, and registers it and its glossary terms in $\mathcal{C}$. \\
\addlinespace[1.5pt]
\texttt{CreateDistribution} & -- &
Defines a probability distribution over a specified domain. \\
\addlinespace[1.5pt]
\texttt{GenValue} & -- &
Generates a value consistent with the supplied description. \\
\addlinespace[1.5pt]
\texttt{HoldConversation} & $\mathcal{D}_{\text{doc}}$ &
Generates a multi-turn stakeholder dialogue and records the decision outcome. \\
\addlinespace[2pt]
\rowcolor{subbox}\multicolumn{3}{@{}l}{\rule{0pt}{2.2ex}\textbf{Stochastic}} \\
\addlinespace[1.5pt]
\texttt{Sample} & -- &
Draws samples from a declared distribution. \\
\addlinespace[2pt]
\rowcolor{subbox}\multicolumn{3}{@{}l}{\rule{0pt}{2.2ex}\textbf{Artifact}} \\
\addlinespace[1.5pt]
\texttt{ApplyTransactions} & $\mathcal{D}_{\text{db}}$ &
Generates and writes a batch of rows from declared per-column generators. \\
\addlinespace[1.5pt]
\texttt{Query} & $\mathcal{D}_{\text{db}}, \mathcal{C}$ &
Runs a query, stores the result as a materialized view, and registers it in $\mathcal{C}$. \\
\addlinespace[1.5pt]
\texttt{AlterSchema} & $\mathcal{D}_{\text{db}}, \mathcal{C}$ &
Snapshots the table, applies a schema modification or migration, and updates $\mathcal{C}$. \\
\addlinespace[1.5pt]
\texttt{ReadDocuments} & -- &
Retrieves unstructured content from $\mathcal{D}_{\text{doc}}$ without modifying state. \\
\addlinespace[2pt]
\rowcolor{subbox}\multicolumn{3}{@{}l}{\rule{0pt}{2.2ex}\textbf{Control}} \\
\addlinespace[1.5pt]
\texttt{Loop} & -- &
Iterates an instruction sequence over a range, a fixed count, or until a predicate holds. \\
\addlinespace[1.5pt]
\texttt{Schedule} & -- &
Schedules an action after a virtual-time delay. \\
\bottomrule
\end{tabular}
\end{table}

\subsection{Simulation Primitives as a DSL}
\label{sec:compilation}

In Stage~2, the compiler translates each approved event type into an \emph{SP program}, which is a sequence of operators drawn from the fixed set $\mathcal{S}$ of twelve \emph{simulation primitives} (\sps) listed in Table~\ref{tab:simulation_primitives}. These twelve primitives fall into four categories, namely \emph{generative}, \emph{stochastic}, \emph{artifact}, and \emph{control} primitives.
Because a recurring event type executes across multiple calendar instances and its SP program can iterate via \texttt{Loop}, a single \emph{operator} in a compiled SP program can produce many runtime \emph{operator invocations}.

\system restricts all LLM-driven state changes to the primitive set $\mathcal{S}$. We designed these twelve primitives to balance execution reliability against program length. Fine-grained primitives are simple enough for an LLM to execute without hallucinating schemas or corrupting state, while coarse-grained primitives keep SP programs concise and readable. Each primitive performs a single bounded action, such as creating a table schema, writing a document, or inserting a batch of rows from column generators, and declares all of its input and output operands explicitly. Explicit operands allow the compiler to verify cross-event references statically before execution and allow the execution engine to record exact provenance in $\mathcal{L}_{\text{log}}$. In this way, $\mathcal{S}$ defines the \emph{boundary of trust} between human specifications and LLM execution. Above the boundary, human-approved event types, compiler checks, and scheduling govern the simulation structure. Below the boundary, the LLM executes individual primitives. Developers can also extend $\mathcal{S}$ with new primitives to support additional data modalities, such as streaming event logs.

\smallskip
\noindent\textbf{What compiled programs guarantee.}
The compiler enforces four properties before handing SP programs to the scheduler.
(1) \emph{Cross-event consistency.} When SP programs of different event types refer to the same table, document, or distribution, they resolve to the same shared artifact. When a disruption event migrates a table via \texttt{AlterSchema}, subsequent operators reference the updated table while historical rows written before the migration retain \texttt{NULL} in newly added columns.
(2) \emph{Referential integrity.} Every foreign-key column in \texttt{ApplyTransactions} draws its values from existing keys in its parent table.
(3) \emph{One program per event type.} Each recurring event type compiles into a single parameterized SP program whose instance-specific values (such as the current date window) bind at run time, and whose per-instance outputs carry distinct dated names so that later instances do not overwrite earlier artifacts.
(4) \emph{Program validity.} Every SP program defines each symbol before use, passes arguments that match primitive signatures, references only existing tables and columns, and implements all operations and outputs required by its event type description.

\subsection{Scheduling and Execution}
\label{sec:sched_exec}
Stage~3 comprises the \emph{scheduler}, which places event instances on the virtual calendar, and the \emph{execution engine}, which runs their SP programs to materialize the data estate and the simulation log.

\subsubsection{Scheduling}
\label{sec:scheduling}

The scheduler takes the approved event types, the compiled SP programs with their data dependencies, and the user's natural-language \emph{simulation instructions} (for example, ``simulate 14 days and upgrade the menu on Day~3''). These instructions specify the \emph{simulation horizon} (start and end dates), recurrence frequencies, and specific dates for disruption events. The scheduler outputs the \emph{calendar} $\mathcal{T}_{\text{sched}}$, which is a set of scheduled event instances in virtual time. Each scheduled instance specifies its event type, start and finish timestamps, and runtime parameter bindings.

A valid calendar satisfies every declared temporal constraint, every data dependency exposed during compilation (ensuring that the event creating an artifact runs before any event that reads it), the recurrence rule of each event type, and the user's simulation instructions. Note that Stage~1 declares temporal constraints between event \emph{types}, whereas the calendar schedules individual \emph{instances}. A type-level constraint applies to the overall time span from the first instance to the last instance of a recurring event type. For example, \textbf{within}(\texttt{Au\-dit},\allowbreak \texttt{Run\-Cafe}) requires that all weekly audits fall between the first and last business days of the cafe, since a weekly audit can span beyond one day's operating window.

\subsubsection{Execution}
\label{sec:execution}

The execution engine runs the scheduled event instances in $\mathcal{T}_{\text{sched}}$ in chronological order of their start times. For each event instance, the engine binds the runtime placeholders and advances a virtual clock across \texttt{Loop} iterations, executing the compiled SP program so that generated rows, documents, and dated snapshots carry accurate creation timestamps.

\smallskip
\noindent\textbf{Populating the catalog.}
As operators execute, the execution engine updates the data catalog $\mathcal{C}$. \texttt{CreateSchema} and \texttt{Query} register newly created tables and views in $\mathcal{C}$. \texttt{AlterSchema} registers the pre-migration snapshot table and updates the schema entry of the migrated authoritative table. \texttt{GenDocument} registers each generated document in $\mathcal{C}$ and extracts structured business-glossary entries, such as status-code definitions and metric formulas.

\smallskip
\noindent\textbf{Recording provenance.}
The execution engine records provenance in $\mathcal{L}_{\text{log}}$ so that the benchmark generator can derive reference answers without human annotation. To keep the log compact without storing every generated tuple, the execution engine appends one provenance record per operator invocation, capturing its event instance, virtual timestamp, resolved input and output symbols, and execution status. It logs schema changes, generated documents, and view definitions in full to preserve exact lineage between authoritative tables and historical snapshots, while summarizing bulk row insertions in \texttt{ApplyTransactions} by row count and column-generator rules. In addition, the execution engine binds each event type's benchmark hints to the physical tables, columns, and documents that the event touched. As a result, $\mathcal{L}_{\text{log}}$ records for every artifact in the data estate which event created or modified it, at what virtual time, from which inputs, and under which rule.

\subsection{Provenance-Based Benchmark Generation}
\label{sec:benchmark_gen}
In Stage~4, the benchmark generator derives evaluation tasks from the data estate  $\mathcal{E}$. Each task pairs a natural-language question with an executable \emph{gold program} whose output on $\mathcal{D}_{\text{db}}$ defines the \emph{reference answer}.
Generating reliable tasks from $\mathcal{L}_{\text{log}}$ without human labeling requires establishing two guarantees for both atomic and composite tasks. First, every question needs to be answerable from the observable assets ($\mathcal{D}_{\text{db}}$, $\mathcal{D}_{\text{doc}}$, and $\mathcal{C}$) without access to $\mathcal{L}_{\text{log}}$. Second, every gold program needs to compute the exact answer required by the question and the logged business rules.

\subsubsection{Answerable tasks}
\label{sec:benchmark_ground_answer}

A benchmark question is answerable only if every schema change, table snapshot, or business rule that affects the answer leaves observable evidence in $\mathcal{D}_{\text{db}}$, $\mathcal{D}_{\text{doc}}$, or $\mathcal{C}$.
The benchmark generator constructs \emph{atomic tasks}, each targeting a single data complexity, from benchmark hints and post-hoc detectors over $\mathcal{L}_{\text{log}}$ and $\mathcal{E}$, and retains a candidate only if it passes three \emph{answerability checks}: (1)~observable artifacts in $\mathcal{D}_{\text{db}}$, $\mathcal{D}_{\text{doc}}$, or $\mathcal{C}$ (such as a migration memo or pre-migration \texttt{NULL} values) anchor the target complexity, (2)~the question uses domain terms without disclosing physical table or column names, requiring discovery via $\mathcal{C}$ and $\mathcal{D}_{\text{doc}}$, and (3)~the candidate gold program executes cleanly on $\mathcal{D}_{\text{db}}$ and returns a non-empty, non-\texttt{NULL} result.

\subsubsection{Verified gold programs}

The benchmark generator constructs each gold program using the complete lineage and business rules recorded in $\mathcal{L}_{\text{log}}$. It checks every factual claim used in deriving a gold program, such as a policy exclusion quoted from a document or an intermediate key mapping in the warehouse, directly against $\mathcal{D}_{\text{doc}}$ and $\mathcal{D}_{\text{db}}$. When a natural-language question permits multiple defensible interpretations, the benchmark generator adds explicit constraints to the question (such as the exact output attributes or the date range) so that the reference answer is unambiguous.
Finally, verified atomic tasks provide reusable \emph{verified query fragments}, namely authoritative table sources, complexity-aware filter predicates, and verified measure expressions. Figure~\ref{fig:intro_example} shows two filter fragments, \texttt{milk\_choice = "oat"} and \texttt{amount > 0}. The benchmark generator composes compatible fragments into a single query to form \emph{composite tasks} that require reasoning across multiple complexities at once, ensuring that each composite gold program inherits the verified logic of its constituent fragments.

\smallskip
\noindent\textbf{Ground-truth audit.} The benchmark generator audits every gold program against $\mathcal{L}_{\text{log}}$, independently of the LLMs that derived it, across four criteria. (1)~The answer agrees with the logged write history (e.g., counting an added column covers only post-migration rows). (2)~The gold program reads only authoritative tables and never a deprecated table or frozen snapshot. (3)~Its filters and measures match the logged generation rules. (4)~Adding rows to a frozen copy of a queried table leaves the answer unchanged. The generator discards any violating task and labels remaining tasks as \emph{log-verified} when a criterion confirms the answer, \emph{corroborated} when only the hint agrees, or \emph{unverified} otherwise.

\smallskip
\noindent\textbf{Task validity checks.} Finally, two deterministic \emph{task validity checks} drop malformed or leaky tasks. A task is \emph{well-formed} if its question and gold-program result are non-empty and no duplicate normalized question yields a different answer. A task has \emph{no schema leakage} if its question names no table read by its gold program (exempting single-table baseline tasks, which name their table).

\section{The \system Implementation}
\label{sec:instance}

This section describes our implementation of \system, built with a coding agent against the stage specifications in Section~\ref{sec:system} and the correctness checks in Section~\ref{sec:eval_pipeline}. We developed the code on three \emph{development scenarios}, froze the implementation, and generated four \emph{held-out scenarios} unchanged.

\subsection{Event Synthesizer}
\label{sec:inst_event}

\noindent\textbf{Candidate event synthesis.}
From the scenario description and requested complexity dimensions, the synthesizer proposes routine event types with parameters, recurrence rules, and benchmark hints, comprising a \emph{setup} event for initial schemas, documents, and distributions and \emph{recurring operational} events for transactions and audits. For each requested dimension, it retrieves a trigger pattern from a \emph{complexity library} covering Table~\ref{tab:complexity_dimensions} (e.g., an unbackfilled column launch, rekeyed migration, or versioned cutover for schema drift), instantiates it for the domain, and links its target artifacts.

\smallskip
\noindent\textbf{Temporal constraint linking.}
Next, the synthesizer connects candidate event types into a directed relational graph by mapping each event role to an interval predicate. It links setup events to dependent routine events with \textbf{precedes} or \textbf{meets}, places disruptions and periodic reviews \textbf{within} the span of recurring operational events with an offset window, and connects phased cutovers, launch promotions, and closing reconciliations with \textbf{overlaps}, \textbf{starts}, and \textbf{finishes}.

\smallskip
\noindent\textbf{Self-critique and repair.}
Finally, a self-critique pass checks the event set and temporal constraints against the scenario description, operating hours, and requested complexity dimensions, revising any conflicting or disconnected event types before HITL review.

\subsection{Compiler}
\label{sec:inst_compiler}

\noindent\textbf{Cross-event consistency.}
\label{sec:cross_event_consistency}
The compiler processes event types in topological order of their temporal constraints and maintains a shared \emph{symbol table}.
When an operator creates a table schema, document, or distribution, the compiler binds the artifact to a \emph{typed symbol} in the symbol table, and operators in later routine and disruption event types pass that symbol directly as an operand. Because the symbol table records which event type binds and reads each symbol, the compiler also derives a temporal constraint requiring each creator to run before its readers. For example, when a migration event applies \texttt{AlterSchema}, the compiler registers symbols for both the snapshot and the migrated authoritative table.

\smallskip
\noindent\textbf{Referential integrity.}
In each \texttt{ApplyTransactions} operator, the \texttt{columns} operand declares how the operator populates each column, whether by copying values from a parent table, run an expression over sibling columns via \texttt{Query}, drawing from a distribution via \texttt{Sample}, assigning a literal, or invoking \texttt{GenValue}. The compiler checks these declarations statically and rejects any foreign-key column that uses \texttt{GenValue} or an independent distribution.

\smallskip
\noindent\textbf{Recurring event instances.}
\label{sec:recurring_instances}
For each recurring event type, the compiler inlines fixed parameters into operator operands and compiles instance-varying values (such as loop counts and time windows) into runtime \emph{placeholders}. For per-instance views, snapshots, and reports, the compiler appends a date placeholder to the output name (such as \texttt{audit\_\allowbreak \{\allowbreak date\}}).

\smallskip
\noindent\textbf{Validation and repair.}
Deterministic \emph{static checks} verify that each SP program defines every input symbol before use, matches operand types to primitive signatures, and references only symbols in the symbol table. An LLM \emph{semantic check} compares each SP program against its event type description to verify that all specified operations and outputs appear. When either check reports a defect, the compiler runs one LLM repair pass and keeps the revised program only if it resolves the defect without introducing new errors.

\subsection{Scheduler and Execution Engine}
\label{sec:inst_sched}

\noindent\textbf{LLM proposal and STN projection.}
The scheduler prompts an LLM with the simulation instructions to propose a start time and duration for each event instance, and projects those proposals onto valid schedules using an STN solver~\citep{dechter1991temporal}. Fixing instance start and finish timestamps sequentially in chronological order, the STN solver computes the \emph{feasible range} of the next timestamp from all active constraints and previously fixed timestamps. The scheduler keeps the LLM proposal if it falls within the feasible range, or clamps it to the nearest bound and updates the remaining bounds.

\smallskip
\noindent\textbf{Applying type-level constraints to instances.}
For an event type with recurring instances $A_1,\dots,A_n$, the scheduler orders consecutive instances via $s_{A_k}<s_{A_{k+1}}$ and $f_{A_k}\leq f_{A_n}$ for $1\leq k<n$, reducing the event type's span $[\min_k s_{A_k},\allowbreak \max_k f_{A_k}]$ to $[s_{A_1},f_{A_n}]$ so every type-level interval predicate encodes as linear inequalities over $s_{A_1}$ and $f_{A_n}$. Before execution, the scheduler replays $\mathcal{T}_{\text{sched}}$ against the evolving schema in calendar order and rejects the calendar if any query reads a column before the instance that creates it.

\subsection{Benchmark Generator}
\label{sec:inst_bench}

\noindent\textbf{Atomic tasks from hints and detectors.}
First, the benchmark generator drops benchmark hints whose target artifacts did not materialize during execution, and passes each remaining hint, the schemas and sample rows of touched tables in $\mathcal{D}_{\text{db}}$, and linked documents in $\mathcal{D}_{\text{doc}}$ to an LLM to generate a natural-language question and candidate gold program. Second, the generator runs two kinds of post-hoc detectors over the materialized data estate. \emph{Deterministic detectors} scan $\mathcal{L}_{\text{log}}$ and $\mathcal{D}_{\text{db}}$ for nine structural complexity patterns (schema drift, missing backfills, versioned table splits, partially backfilled splits, stale replicas, duplicate views, document--database contradictions, implicit business rules, and multi-granularity summaries), while \emph{LLM detectors} inspect generated documents and their log slices to propose cross-asset tasks. Candidates passing the answerability checks enter the atomic task pool.

\smallskip
\noindent\textbf{Composite tasks.}
The generator builds composite tasks either by combining compatible filter and measure fragments into a single aggregate query over foreign-key joins, or by nesting verified atomic gold programs inside outer comparison and ranking queries. It re-executes each composite gold program while ablating one fragment at a time to confirm that every constituent complexity affects the output. An LLM then rephrases each composite specification into a natural-language question, which the generator retains only if a second LLM maps it back to the original query specification.

\smallskip
\noindent\textbf{Gold program derivation by multi-agent debate.}
\label{sec:inst_debate}
An iterative \emph{multi-agent debate} among a panel of LLM agents derives and verifies each gold program over $\mathcal{D}_{\text{db}}$, $\mathcal{D}_{\text{doc}}$, and $\mathcal{L}_{\text{log}}$. In each round, proposer agents backed by multiple distinct LLMs independently write a SQL query for the question and check each clause of the candidate gold program against logged table lineage and policy documents.
When the proposers reach different answers, one agent defends the candidate program's table selection, joins, filters, and formulas, while dissenting critics challenge specific clauses by citing the logged events in $\mathcal{L}_{\text{log}}$ (such as a schema cutover or an in-place table update) and propose a revised query. A judge agent weighs the verified evidence from both sides and either keeps the candidate program, adopts a critic's revised query for the next round, or flags the question as ambiguous. The panel runs for up to five rounds and settles when two consecutive rounds reach high-confidence agreement on equivalent gold programs. When the judge finds that a question admits multiple valid readings under the logged rules, the panel adds explicit scope constraints (such as target output columns or time windows) and reruns the derivation.

\section{Evaluation}
\label{sec:eval}

We generate seven data estates and 256 tasks across five domains using \system. We summarize the scenarios and generated data estates, then evaluate pipeline correctness, data estate complexity and benchmark quality, and agent performance across data-context settings, model tiers, and provenance-verified composite tasks.

\subsection{Scenarios and Generated Data Estates}
\label{sec:eval_scenario}

\begin{table}[t]
\centering\small
\caption{The seven \system data estates and benchmarks.}
\label{tab:estates}
\setlength{\tabcolsep}{3pt}
\renewcommand{\arraystretch}{1.08}
\begin{tabular}{@{}>{\raggedright\arraybackslash}p{\dimexpr\textwidth-11.34cm-14\tabcolsep\relax}>{\raggedleft\arraybackslash}p{1.45cm}>{\raggedleft\arraybackslash}p{1.65cm}>{\raggedleft\arraybackslash}p{1.82cm}>{\raggedleft\arraybackslash}p{1.80cm}>{\raggedleft\arraybackslash}p{1.35cm}>{\raggedleft\arraybackslash}p{1.42cm}>{\raggedleft\arraybackslash}p{1.85cm}@{}}
\toprule
 & \mbox{\textbf{Retail-10}} & \mbox{\textbf{Retail-100}} & \mbox{\textbf{Retail-1000}} & \mbox{\textbf{Music store}} & \mbox{\textbf{Banking}} & \mbox{\textbf{Logistics}} & \mbox{\textbf{Insurance}} \\
\midrule
Complexity & SD & SD & All & IBR & BG & \mbox{TR, MG} & \mbox{IBR, MG, SD} \\
Horizon (days) & 14 & 14 & 28 & 60 & 120 & 30 & 120 \\
\midrule
Event types {\footnotesize(R$+$D)} & $11{+}2$ & $27{+}1$ & $59{+}31$ & $5{+}5$ & $9{+}3$ & $8{+}5$ & $7{+}4$ \\
Operators & 59 & 299 & 1.3K & 90 & 91 & 86 & 92 \\
Event instances & 65 & 153 & 834 & 147 & 152 & 177 & 585 \\
Operator invocations & 280 & 1.7K & 15.7K & 545 & 3.2K & 1.2K & 4.5K \\
\midrule
Total tables & 33 & 272 & 1.4K & 11 & 25 & 60 & 126 \\
Columns & 259 & 707 & 6.9K & 64 & 125 & 609 & 863 \\
Rows & 12.7K & 30.5K & 1.5M & 5.5K & 22.6K & 14.0K & 130.1K \\
Documents & 11 & 104 & 1.4K & 5 & 7 & 26 & 18 \\
Catalog entries & 189 & 2.0K & 6.8K & 69 & 173 & 472 & 379 \\
\quad glossary terms & 146 & 1.6K & 5.0K & 53 & 144 & 405 & 240 \\
\midrule
Tasks & 37 & 31 & 62 & 28 & 26 & 19 & 53 \\
\quad atomic / composite & 23/14 & 23/8 & 36/26 & 17/11 & 18/8 & 14/5 & 24/29 \\
\quad doc-required & 32 & 25 & 48 & 24 & 18 & 13 & 45 \\
\midrule
Runtime (min) & & & & & & & \\
\quad compilation & 22.5 & 66.1 & 353.3 & 11.8 & 17.5 & 32.7 & 20.8 \\
\quad execution & 4.5 & 19.5 & 340.8 & 10.3 & 3.6 & 11.9 & 42.9 \\
\midrule
LLM calls & & & & & & & \\
\quad compilation & 17 & 35 & 110 & 10 & 12 & 17 & 13 \\
\quad execution & 79 & 450 & 2.8K & 137 & 1.3K & 301 & 2.0K \\
\bottomrule
\end{tabular}
\\[3pt]
\parbox{\textwidth}{\footnotesize Complexity dimensions are business glossary (BG), implicit business rules (IBR), multi-granularity (MG), table redundancy (TR), and schema drift (SD). R$+$D denotes routine $+$ disruption event types. Retail-10, Retail-100, and Retail-1000 indicate their number of operational tables.}
\end{table}

We generate six data estates with Gemini~4 Argon and generate Retail-1000 with Gemini~3.8 Flash, using Retail-10, Banking, and Logistics as development scenarios and generating the remaining four held-out scenarios after freezing the implementation. Table~\ref{tab:estates} summarizes the statistics of the seven data estates and benchmarks, and Appendix~\ref{app:scenarios} details each scenario.
\textbf{Music store} builds on the public Chinook schema\footnote{\url{https://github.com/lerocha/chinook-database}} with policy memos for QA canary checkouts, in-place anonymized closed accounts, and off-database flash-sale refunds. \textbf{Banking} builds on the Czech bank schema from BIRD \texttt{financial}~\citep{li2023can}, pairing a Czech codebook of transaction codes, KPIs, and fee schedules with a Day~45 steelworks closure that pushes loans into arrears. \textbf{Logistics} adapts the ERP payment-to-invoice reconciliation task of \citet{bodensohn2025unveiling} with a frozen Day~12 auditor invoice snapshot, a bank-transaction replica that stops refreshing on Day~18, and daily and monthly treasury summaries. \textbf{Retail-10}, \textbf{Retail-100}, and \textbf{Retail-1000} model a specialty coffee business at 10, 100, and 1K operational tables with a shared counter-sales core and an unbackfilled \texttt{milk\_choice} migration, scaling from one shop (\textbf{Retail-10}) to 14 back-office subsystems (\textbf{Retail-100}) and a 20-store chain with an acquired competitor, vendor landing zones, and an analytics warehouse (\textbf{Retail-1000}). \textbf{Insurance} models a property-and-casualty insurer with biweekly QA canary claims, multi-invoice lockbox lump sums, and an unbackfilled Day~48 telematics rating migration.

\smallskip
\noindent \textbf{Data estate composition.}
Across the seven data estates, \texttt{CreateSchema} and \texttt{AlterSchema} produce 1,176 base tables, while recurring \texttt{Query} operators materialize report snapshots (41.7\% of tables per data estate on average, and 52.9\% including views). Reference and transaction tables form 67.3\% of base tables, and transaction tables hold over 90\% of rows in five data estates. Retail-1000 features a multi-layer warehouse architecture (357 operational source tables, 346 vendor and acquired-system feeds, 235 staging and warehouse tables, and 66 frozen or sandbox copies), whereas the six smaller data estates hold operational tables plus at most two frozen snapshots from migrations or audits. Across document stores, 1,562 documents instantiate 228 templates spanning dated reports, policies, data dictionaries, migration notes, and transcripts. Appendix~\ref{app:composition} and Table~\ref{tab:estate_composition} report the full table and document breakdown across all seven data estates.

\subsection{Pipeline Correctness}
\label{sec:eval_pipeline}

\begin{table}[t]
\centering\small
\caption{Pipeline correctness on the development and held-out scenarios. The LLM judge evaluates the \ul{underlined} error types, and the rule engine evaluates the rest. Parentheses give the counting unit. Each cell reports affected over total units (and the error rate when nonzero). Shading follows the error rate (\errbar).}
\label{tab:pipeline_eval}
\setlength{\tabcolsep}{3.5pt}
\renewcommand{\arraystretch}{1.15}
\resizebox{\textwidth}{!}{%
\begin{tabular}{@{}lrrrrrrr}
\toprule
 & \multicolumn{3}{c}{\textbf{Development}} & \multicolumn{4}{c}{\textbf{Held-out}} \\
\cmidrule(lr){2-4}\cmidrule(l){5-8}
\textbf{Error type} & \textbf{Retail-10} & \textbf{Banking} & \textbf{Logistics} & \textbf{Retail-100} & \textbf{Retail-1000} & \textbf{Music store} & \textbf{Insurance} \\
\midrule
\multicolumn{8}{@{}l}{\textbf{Compiled SP programs}}\\
\ul{Cross-column/row dependency violation} (dependencies) & \qcell{fiderr!30!qmid}{3 / 105 (0.03)} & \qcell{fiderr!45!qmid}{8 / 104 (0.08)} & \qcell{fiderr!30!qmid}{4 / 129 (0.03)} & \qcell{fiderr!45!qmid}{18 / 253 (0.07)} & \qcell{fiderr!60!qmid}{186 / 1.1K (0.17)} & \qcell{fiderr!20!qmid}{1 / 75 (0.01)} & \qcell{fiderr!50!qmid}{14 / 109 (0.13)} \\
\ul{Missing required output} (required outputs) & \qcell{qhigh}{0 / 33} & \qcell{qhigh}{0 / 32} & \qcell{qhigh}{0 / 53} & \qcell{fiderr!15!qmid}{1 / 236 (0.00)} & \qcell{fiderr!25!qmid}{25 / 1.5K (0.02)} & \qcell{qhigh}{0 / 23} & \qcell{fiderr!40!qmid}{3 / 52 (0.06)} \\
\ul{Missing stochastic sampling} (value-producing operators) & \qcell{fiderr!35!qmid}{2 / 47 (0.04)} & \qcell{fiderr!25!qmid}{2 / 85 (0.02)} & \qcell{fiderr!30!qmid}{2 / 74 (0.03)} & \qcell{fiderr!20!qmid}{3 / 257 (0.01)} & \qcell{fiderr!25!qmid}{19 / 1.0K (0.02)} & \qcell{qhigh}{0 / 84} & \qcell{qhigh}{0 / 83} \\
Unbound variable (variables) & \qcell{qhigh}{0 / 60} & \qcell{qhigh}{0 / 91} & \qcell{qhigh}{0 / 86} & \qcell{qhigh}{0 / 300} & \qcell{qhigh}{0 / 1.3K} & \qcell{qhigh}{0 / 90} & \qcell{qhigh}{0 / 93} \\
Unintended output overwrite (operators) & \qcell{qhigh}{0 / 59} & \qcell{qhigh}{0 / 91} & \qcell{qhigh}{0 / 86} & \qcell{qhigh}{0 / 299} & \qcell{qhigh}{0 / 1.3K} & \qcell{qhigh}{0 / 90} & \qcell{qhigh}{0 / 92} \\
Foreign key violation (tables) & \qcell{qhigh}{0 / 5} & \qcell{qhigh}{0 / 8} & \qcell{qhigh}{0 / 15} & \qcell{qhigh}{0 / 69} & \qcell{fiderr!15!qmid}{2 / 537 (0.00)\textsuperscript{a}} & \qcell{qhigh}{0 / 7} & \qcell{qhigh}{0 / 15} \\
\addlinespace\multicolumn{8}{@{}l}{\textbf{Simulation primitives}}\\
Undeclared write (operators) & \qcell{qhigh}{0 / 59} & \qcell{qhigh}{0 / 91} & \qcell{qhigh}{0 / 86} & \qcell{qhigh}{0 / 299} & \qcell{qhigh}{0 / 1.3K} & \qcell{qhigh}{0 / 90} & \qcell{qhigh}{0 / 92} \\
\ul{Operator contract violation} (operators) & \qcell{fiderr!25!qmid}{1 / 59 (0.02)} & \qcell{fiderr!30!qmid}{3 / 91 (0.03)} & \qcell{fiderr!40!qmid}{6 / 86 (0.07)} & \qcell{fiderr!35!qmid}{13 / 299 (0.04)} & \qcell{fiderr!45!qmid}{93 / 1.3K (0.07)} & \qcell{qhigh}{0 / 90} & \qcell{fiderr!30!qmid}{3 / 92 (0.03)} \\
Placeholder value (text columns) & \qcell{qhigh}{0 / 207} & \qcell{qhigh}{0 / 53} & \qcell{qhigh}{0 / 412} & \qcell{qhigh}{0 / 332} & \qcell{fiderr!15!qmid}{5 / 3.0K (0.00)} & \qcell{qhigh}{0 / 37} & \qcell{qhigh}{0 / 704} \\
\addlinespace\multicolumn{8}{@{}l}{\textbf{Calendar}}\\
Unscheduled event type (event types) & \qcell{qhigh}{0 / 13} & \qcell{qhigh}{0 / 12} & \qcell{qhigh}{0 / 13} & \qcell{qhigh}{0 / 28} & \qcell{qhigh}{0 / 90} & \qcell{qhigh}{0 / 10} & \qcell{qhigh}{0 / 11} \\
Recurrence violation (event types) & \qcell{qhigh}{0 / 13} & \qcell{qhigh}{0 / 12} & \qcell{qhigh}{0 / 13} & \qcell{qhigh}{0 / 28} & \qcell{qhigh}{0 / 90} & \qcell{qhigh}{0 / 10} & \qcell{qhigh}{0 / 11} \\
\ul{Instruction mismatch} (event types) & \qcell{fiderr!45!qmid}{1 / 13 (0.08)} & \qcell{qhigh}{0 / 12} & \qcell{qhigh}{0 / 13} & \qcell{qhigh}{0 / 28} & \qcell{fiderr!20!qmid}{1 / 90 (0.01)} & \qcell{qhigh}{0 / 10} & \qcell{qhigh}{0 / 11} \\
\addlinespace\multicolumn{8}{@{}l}{\textbf{Execution}}\\
Database execution failure (invocations) & \qcell{qhigh}{0 / 280} & \qcell{qhigh}{0 / 3.2K} & \qcell{qhigh}{0 / 1.2K} & \qcell{qhigh}{0 / 1.7K} & \qcell{fiderr!15!qmid}{33 / 15.7K (0.00)\textsuperscript{b}} & \qcell{qhigh}{0 / 545} & \qcell{qhigh}{0 / 4.5K} \\
Unexpectedly empty table (write targets) & \qcell{qhigh}{0 / 10} & \qcell{qhigh}{0 / 9} & \qcell{qhigh}{0 / 22} & \qcell{qhigh}{0 / 100} & \qcell{fiderr!30!qmid}{34 / 1.0K (0.03)\textsuperscript{c}} & \qcell{qhigh}{0 / 11} & \qcell{qhigh}{0 / 17} \\
\bottomrule
\end{tabular}%
}
\\[3pt]
\parbox{\textwidth}{\footnotesize \textsuperscript{a}The queries that write both tables look up a foreign key in an empty reference table and fall back to a default key that has no parent row. \textsuperscript{b}Writes fail when no parent row matches a foreign-key filter. One daily write fails on all 28 days (28 of the 33 failures). \textsuperscript{c}That failed daily write targets one of the 34 tables. Warehouse refresh queries that return no rows, because they read empty upstream tables or apply mismatched filters, target the other 33.}
\end{table}

Pipeline correctness tests whether \system delivers the guarantees of each stage. Table~\ref{tab:pipeline_eval} defines 14 error types in four groups, and Appendix~\ref{app:pipeline_checks} and Table~\ref{tab:defect_taxonomy} specify their detection criteria and counting units. Six test compiled SP programs for validity, single-program parameterization, and referential integrity, three test whether primitives stay within declared operands below the boundary of trust, three test calendar recurrence and instruction compliance, and two test runtime database execution and non-empty write targets. We do not score Stage~1 because users approve the event set in HITL review.

A deterministic \emph{rule engine} checks the nine deterministic error types over SP programs, $\mathcal{T}_{\text{sched}}$, $\mathcal{L}_{\text{log}}$, and $\mathcal{D}_{\text{db}}$. An \emph{LLM judge} (Claude Opus~5.5) evaluates the five semantic error types across all the event types. Given each event specification, compiled SP program, and runtime evidence in a single call, the judge cites candidate defects with data evidence and retains only defects that alter analytical query answers or break dependencies named in the specification. Each error type counts at the unit where it arises. For cross-column/row dependency violations and missing required outputs, an LLM enumerates the required dependencies and outputs from each specification and matches judge findings against them.

\smallskip
\noindent \textbf{Structural correctness holds.}
On the six smaller data estates, all nine deterministic checks pass across both development and held-out scenarios.
On Retail-1000, every static program and calendar check passes, and the remaining runtime errors stem from a few mismatched query filters that cascade across time and warehouse layers. When a query in a daily event applies a mismatched filter, it fails on every scheduled day, and when an upstream refresh query returns zero rows, downstream warehouse tables and foreign-key lookups that read from that empty table also remain empty.

\smallskip
\noindent \textbf{Semantic errors arise in compilation.}
Semantic errors appear when an SP program coordinates dependencies across multiple columns, rows, or tables.
Across value-producing operators, 1.7\% replace a stochastic draw with a constant. Operator contract violations affect 3.6\% of operators on the six smaller data estates and 7.1\% on Retail-1000, mostly when generated documents state figures, status codes, or column descriptions that contradict the schema or data. Calendar instruction mismatches occur once on Retail-10 and once on Retail-1000.
\emph{Cross-column/row dependency violations} affect 6.2\% of specified dependencies on the six smaller data estates and 12.7\% across all seven, occurring when compiled programs sample dependent columns independently, flatten one-to-many cardinalities, substitute fixed or aggregate values for per-row formulas, read unrefreshed state, or apply mismatched filters.
\emph{Missing required output} errors affect 0.9\% of required outputs on the six smaller data estates and 1.5\% overall, concentrating on Retail-1000 where individual events coordinate many warehouse tables and data marts.
Reference answers remain unaffected by semantic compilation errors because \system executes every gold program on the materialized database $\mathcal{D}_{\text{db}}$ and audits every task against $\mathcal{L}_{\text{log}}$.

\subsection{Data Estate Complexity and Benchmark Quality}
\label{sec:eval_complexity}

\noindent \textbf{Data estate complexity.}
Across $\mathcal{D}_{\text{db}}$, redundant table copies, including backups, vendor mirrors, staging tables, and reporting replicas, coexist with authoritative tables on 68\% of Retail-1000 tasks and 63\% of Logistics tasks, compared with 0\% to 38\% on the other five data estates. On Retail-1000, BM25 search over table names, columns, and catalog descriptions retrieves all gold tables of a task within the top 50 results on 55\% of tasks. Across $\mathcal{D}_{\text{doc}}$, 63\% of document-dependent tasks draw all required business facts (such as status codes, KPI formulas, cutover dates, and authoritative table names) from a single document, while 33\% combine facts across two or more documents. Moreover, 64\% of tasks encounter a superseded statement of a required fact elsewhere in $\mathcal{D}_{\text{doc}}$ or $\mathcal{C}$'s glossary. Retail-1000 exhibits the highest document dispersion, where 50\% of tasks combine multiple documents and 20\% of required facts appear implicitly through examples or report figures. Once an agent retrieves the relevant document, applying its rules is direct because 91\% of required facts name the target physical table, column, or code value and 93\% of documents include explicit SQL fragments such as \texttt{WHERE} predicates or \texttt{SUM} aggregates.

\smallskip
\noindent \textbf{Task complexity and SQL diversity.}
Of the 256 generated tasks, 80.1\% (205 tasks) combine SQL queries with evidence from documents or the catalog, and 39.5\% (101 tasks) are multi-complexity composite tasks. Every benchmark covers all requested complexity dimensions. Requested disruptions produce 60.1\% of the 198 complexity-bearing tasks (97.9\% on Retail-1000), while routine events contribute 39.9\% via status-code filters, discount policies, and dated snapshots. On 68.0\% of single-measure numeric tasks, no report states the answer so agents need to query $\mathcal{D}_{\text{db}}$. Overall, 91.8\% of gold programs have a unique syntax skeleton, and on average across benchmarks, 81.0\% use grouping and aggregation, 57.4\% use nested subqueries, and 59.3\% join multiple tables. Appendix~\ref{app:benchmark_checks} and Table~\ref{tab:benchmark_quality} detail the benchmark quality checks and per-estate SQL diversity metrics.

\begin{table}[t]
\centering\small
\caption{BIRD \texttt{financial} versus the \system Banking benchmark on the same Czech bank schema.}
\label{tab:bird_vs_banking}
\setlength{\tabcolsep}{6pt}
\renewcommand{\arraystretch}{1.15}
\begin{tabular}{@{}>{\raggedright\arraybackslash}p{2.5cm}>{\raggedright\arraybackslash}p{6.5cm}>{\raggedright\arraybackslash}p{\dimexpr\textwidth-9.0cm-4\tabcolsep\relax}@{}}
\toprule
 & \textbf{BIRD \texttt{financial}} & \textbf{\system Banking} \\
\midrule
\textbf{Data} & Static snapshot of a Czech bank (8 tables) & 120 simulated days (9 base tables, 12 dated snapshots, 4 views) \\
\textbf{History} & None & Simulation log of 152 event instances \\
\textbf{Knowledge} & Column descriptions and per-question external-knowledge evidence & 7 event-generated documents, no column descriptions \\
\textbf{Catalog} & -- & 144 glossary terms linked to source documents \\
\textbf{Questions} & Human-written & Generated from benchmark hints and detectors \\
\textbf{Gold program} & Human-written & Log-derived and debate-verified \\
\textbf{Access} & Annotator supplies evidence in the prompt & Agent retrieves the term, document, and rule \\
\bottomrule
\end{tabular}
\end{table}

\subsection{Case Study on BIRD \texttt{financial}}
\label{sec:eval_bird}

We compare the \system Banking benchmark with BIRD~\citep{li2023can}, whose \texttt{financial} database shares the Czech bank schema. Table~\ref{tab:bird_vs_banking} contrasts the two benchmarks. BIRD attaches human-written \emph{external-knowledge evidence} to each prompt (such as the meaning of loan status \texttt{\textquotesingle{}D\textquotesingle{}}). In \system Banking, business events embed domain knowledge inside policy documents and codebooks, the catalog links glossary terms to source documents, and questions use domain terms without inlining definitions.

\definecolor{glossfg}{RGB}{0, 118, 112}
\definecolor{docfg}{RGB}{176, 82, 0}
\newcommand{\fromgloss}[1]{\textcolor{glossfg}{#1}}
\newcommand{\fromdoc}[1]{\textcolor{docfg}{#1}}
\begin{tcolorbox}[colback=gray!4,colframe=gray!50,boxrule=0.4pt,left=3pt,right=3pt,top=2pt,bottom=2pt,fontupper=\footnotesize]
\textbf{Example task.} \emph{``Rank our districts by Net Deposit Inflow across the full transaction history we hold.''}\\[2pt]
\textbf{Gold program derivation.}\\
\circled{1} \textbf{\fromgloss{Glossary.}} A catalog search returns \fromgloss{\emph{Net Deposit Inflow (NDI)}} and its source, the \fromgloss{\emph{Core Banking Codebook}}.\\
\circled{2} \textbf{\fromdoc{Document.}} The codebook defines NDI as \fromdoc{credits (\texttt{PRIJEM}) minus debits (\texttt{VYDAJ}, \texttt{VYBER}), excluding loan flows (\texttt{k\_symbol = "UVER"}) and interest (\texttt{"UROK"})}, and notes that \fromdoc{regular transfers have a \texttt{NULL} \texttt{k\_symbol}}.\\
\circled{3} \textbf{SQL.} The gold program applies these \fromdoc{exclusions} with \fromdoc{\texttt{COALESCE(\allowbreak k\_symbol,"")}} to retain \texttt{NULL}-coded rows.
\end{tcolorbox}

Beyond glossary lookup, Banking tasks require \emph{event-grounded filtering} (excluding relocation withdrawals during the Day~45 steelworks closure documented in an operational bulletin), \emph{snapshot selection} (choosing the authoritative table over 12 month-end snapshots without double-counting balances across months), and \emph{code disambiguation} (separating \texttt{UVER} disbursements from repayments by direction). Without documents in setting S1, agents miss these rules, whereas single-shot retrieval in S2 raises Banking accuracy by 8 to 12 percentage points and multi-step agentic search in S3 raises it by 23 to 28 points across all three models.

\begin{table}[t]
\centering\small
\caption{Mean execution accuracy ($\pm$ standard error, \%) across four data-context settings on all 256 tasks over three runs. Bold marks each model's best setting, and \ul{\textbf{bold underline}} marks each column's best result.}
\label{tab:agent_eval}
\setlength{\tabcolsep}{4.5pt}
\renewcommand{\arraystretch}{1.12}
\newcommand{\sd}[1]{{\scriptsize$\pm$#1}}
\resizebox{\textwidth}{!}{%
\begin{tabular}{@{}lrrrrrrrr@{}}
\toprule
 & \textbf{Retail-10} & \textbf{Retail-100} & \textbf{Retail-1000} & \textbf{Music store} & \textbf{Banking} & \textbf{Logistics} & \textbf{Insurance} & \textbf{All} \\
\#Tasks & 37 & 31 & 62 & 28 & 26 & 19 & 53 & 256 \\
\midrule
\multicolumn{9}{@{}l}{\textbf{Gemma~4 26B-A4B} }\\
S1 {\footnotesize (Schema only)} & 27.9\sd{7.1} & 64.5\sd{8.0} & 22.0\sd{4.9} & 16.7\sd{7.0} & 37.2\sd{9.5} & \textbf{66.7}\sd{9.9} & 27.0\sd{6.0} & 33.3\sd{2.8} \\
S2 {\footnotesize (Schema + top-1 document)} & 43.2\sd{7.7} & \textbf{67.7}\sd{8.2} & 26.9\sd{5.3} & 28.6\sd{8.7} & 44.9\sd{9.2} & 47.4\sd{11.8} & 49.1\sd{6.6} & 42.3\sd{3.0} \\
S3 {\footnotesize (Agent + search tools)} & \textbf{63.1}\sd{7.7} & 64.5\sd{8.0} & \textbf{34.9}\sd{5.5} & \textbf{63.1}\sd{8.1} & \textbf{64.1}\sd{7.6} & 64.9\sd{9.7} & \textbf{52.8}\sd{6.3} & \textbf{54.6}\sd{2.8} \\
S4 {\footnotesize (Agent + files and CLI)} & 59.5\sd{7.7} & 60.2\sd{7.9} & 23.7\sd{4.7} & 50.0\sd{7.4} & 56.4\sd{7.6} & 59.6\sd{10.7} & 47.8\sd{6.3} & 47.1\sd{2.8} \\
\addlinespace[3pt]
\multicolumn{9}{@{}l}{\textbf{DeepSeek V4.1 Flash} }\\
S1 {\footnotesize (Schema only)} & 42.3\sd{7.7} & 73.1\sd{7.0} & 30.6\sd{5.4} & 14.3\sd{6.7} & 57.7\sd{9.0} & \textbf{77.2}\sd{8.8} & 30.8\sd{6.1} & 41.9\sd{2.9} \\
S2 {\footnotesize (Schema + top-1 document)} & 64.9\sd{7.3} & \textbf{75.3}\sd{7.1} & 39.8\sd{5.7} & 33.3\sd{8.7} & 69.2\sd{8.0} & 63.2\sd{10.2} & 61.6\sd{6.1} & 56.2\sd{2.8} \\
S3 {\footnotesize (Agent + search tools)} & \textbf{78.4}\sd{5.9} & 69.9\sd{7.5} & 71.0\sd{4.5} & 77.4\sd{6.9} & 85.9\sd{6.2} & \textbf{77.2}\sd{7.2} & \textbf{80.5}\sd{4.9} & 76.6\sd{2.2} \\
S4 {\footnotesize (Agent + files and CLI)} & 76.6\sd{6.0} & 72.0\sd{6.6} & \textbf{71.5}\sd{4.7} & \textbf{85.7}\sd{5.3} & \ul{\textbf{91.0}}\sd{5.1} & 75.4\sd{7.6} & 79.2\sd{4.9} & \textbf{77.7}\sd{2.2} \\
\addlinespace[3pt]
\multicolumn{9}{@{}l}{\textbf{Claude Opus 5.5} }\\
S1 {\footnotesize (Schema only)} & 53.2\sd{7.6} & 77.4\sd{7.6} & 54.3\sd{5.7} & 15.5\sd{6.7} & 65.4\sd{9.5} & 82.5\sd{7.8} & 47.8\sd{6.4} & 54.6\sd{2.9} \\
S2 {\footnotesize (Schema + top-1 document)} & 83.8\sd{6.1} & 79.6\sd{7.0} & 68.3\sd{5.4} & 42.9\sd{8.7} & 76.9\sd{8.4} & 80.7\sd{8.6} & 69.8\sd{5.9} & 71.2\sd{2.7} \\
S3 {\footnotesize (Agent + search tools)} & 90.1\sd{4.4} & \ul{\textbf{84.9}}\sd{5.5} & \ul{\textbf{89.2}}\sd{3.7} & 78.6\sd{7.1} & \textbf{88.5}\sd{5.5} & \ul{\textbf{86.0}}\sd{6.9} & 85.5\sd{4.4} & 86.6\sd{1.9} \\
S4 {\footnotesize (Agent + files and CLI)} & \ul{\textbf{94.6}}\sd{3.8} & 81.7\sd{6.5} & 87.6\sd{3.5} & \ul{\textbf{94.0}}\sd{3.0} & 87.2\sd{5.9} & \ul{\textbf{86.0}}\sd{7.4} & \ul{\textbf{89.3}}\sd{3.7} & \ul{\textbf{88.8}}\sd{1.7} \\
\bottomrule
\end{tabular}%
}
\end{table}

\subsection{Agent Performance}
\label{sec:eval_agents}

We evaluate data agents on the \system benchmarks to answer three questions. First, does the benchmark separate data-context settings and model tiers within each setting? Second, when state-of-the-art agents fail, do those failures stem from gold-program errors, format inconsistencies, or genuine reasoning mistakes? Third, on composite tasks, does $\mathcal{L}_{\text{log}}$ both verify reference answers and challenge state-of-the-art agents?

Table~\ref{tab:agent_eval} reports the execution accuracy of three reasoning-enabled models across four data-context settings over three runs per task.
We evaluate \textbf{Gemma~4 26B-A4B} (Gemma, compact open-weight MoE with 4B active parameters), \textbf{DeepSeek V4.1 Flash} (DeepSeek, 749B open-weight MoE), and \textbf{Claude Opus 5.5} (Opus, proprietary frontier model). Each setting expands how the model accesses the data estate.
\begin{itemize}[leftmargin=.45cm,itemsep=1pt,topsep=2pt]
\item \textbf{S1 (Single-shot + Schema only).} A single text-to-SQL prompt with the question and the full DDL of $\mathcal{D}_{\text{db}}$, plus one retry on a SQL execution error, testing whether physical schemas suffice alone. On Retail-1000, all tables and views take 71K to 157K tokens across the three tokenizers and fit in context.
\item \textbf{S2 (Single-shot + Schema and top-1 document).} S1 plus the top BM25-retrieved document from $\mathcal{D}_{\text{doc}}$, testing the effect of single-shot document retrieval.
\item \textbf{S3 (Agent + search tools).} A multi-step catalog-backed agent with seven tools (no CLI or file access) to list tables, BM25-search tables, documents, and the glossary in $\mathcal{C}$, read schemas and documents, and run read-only SQL queries.
\item \textbf{S4 (Agent + files and CLI).} A multi-step CLI agent in a sandbox exposing $\mathcal{D}_{\text{db}}$ and the glossary in $\mathcal{C}$ as SQLite files and $\mathcal{D}_{\text{doc}}$ as text files, with file tools and a read-only \texttt{sqlite3} CLI.
\end{itemize}

S3 and S4 expose the same assets through search tools versus raw files, isolating the agent interface when tasks exceed single-shot retrieval. The evaluator runs each submitted query on $\mathcal{D}_{\text{db}}$ and grades it correct when every reference row matches a distinct output row on all reference values, ignoring extra columns, ordering, and float rounding.

\smallskip
\noindent \textbf{Separating settings and model tiers.}
Table~\ref{tab:agent_eval} shows a consistent ordering across model tiers and data-context settings. Opus leads DeepSeek and DeepSeek leads Gemma in every setting, and the accuracy gap between the frontier model and the open-weight models widens from single-shot prompting (S1) to multi-step agentic search (S3 and S4) as stronger models plan multi-hop retrieval and recover from SQL errors.
Across settings, schema-only prompting (S1) fails whenever business rules live outside DDLs. Adding a single retrieved document (S2) substantially improves accuracy on data estates where one policy memo defines the target status codes or exclusions (such as Retail-10 and Insurance), whereas data estates with multi-document rules or large schemas (such as Music store and Retail-1000) require multi-step agentic search (S3 and S4). The largest multi-step gains appear on Retail-1000, where agents navigate thousands of tables and documents and where open-weight models lag behind Opus.
Comparing the two agentic interfaces, structured search tools (S3) and raw file access (S4) achieve similar overall accuracy on both Opus and DeepSeek. Direct file inspection in S4 performs well on small document stores such as Music store, whereas paged search tools in S3 raise Gemma's accuracy across large data estates by cutting unanswered runs from 8.3\% to 1.2\%. Removing the pre-built glossary from S4 changes overall accuracy by less than 3 points across all models, as Opus and DeepSeek recover the same definitions from the source documents.

\smallskip
\noindent \textbf{Cost of data-context settings.}
We measure cost by the input tokens, output tokens, and model calls per task run. Input tokens dominate, since output tokens, including reasoning, stay below 9K per run in every setting. Single-shot prompts grow with the schema. An S1 prompt for Opus takes 1.5K tokens on Music store and 157K tokens on Retail-1000, so schema-only prompting stops fitting in context as warehouses grow. One retrieved document in S2 adds at most 1.5K input tokens per run yet raises accuracy by 9 to 17 points, the cheapest gain across settings.
Multi-step agents trade tokens for accuracy. S3 agents make 9 to 11 model calls per run and resend the growing conversation in each call, which raises input tokens over S1 by 1.8 times for Opus, 4.4 times for Gemma, and 12 times for DeepSeek. Stronger models explore more efficiently. DeepSeek issues 26.5 tool calls and 276K input tokens per S3 run, compared with 10.2 tool calls and 88K tokens for Opus, even though DeepSeek's tokenizer counts about half as many tokens as Claude's for the same text. Normalized by accuracy, Opus needs 102K input tokens per correct S3 answer, against 266K for Gemma and 360K for DeepSeek.
File access in S4 raises cost without improving accuracy. Shell commands and file reads often return whole documents and full query results, whereas search tools return pages of ten hits, so each Opus call carries 25K input tokens in S4 compared with 10K in S3. Per correct answer, S4 consumes 226K input tokens for Opus and 934K for DeepSeek, 2.2 and 2.6 times their S3 cost.

\smallskip
\noindent \textbf{Auditing reference answers and agent failures.}
In S1, most failures happen because the prompt provides only the database schema and omits the business rules in $\mathcal{D}_{\text{doc}}$. Without those documents, over a third of S1 runs fail.
The S4 failures come from two sources, namely format inconsistencies between the gold output and the agent's prediction, and genuine reasoning failures where the agent fails to find or apply the required domain knowledge.

First, format issues cause 56\% of Opus's failed runs and 61\% of DeepSeek's failed runs in S4, where the agent computes the right numbers but formats its output differently from the reference answer (such as returning entity names in place of IDs, pivoting rows into columns, or omitting an extra column in the gold program). Counting these predictions as correct raises S4 accuracy to 94\% for Opus and 90\% for DeepSeek. In particular, on the three tasks where every model fails, a manual audit shows that 53 out of 54 failed runs compute the correct reference answers and fail only on format differences.
Because \system generates tasks and gold programs automatically without human SQL annotation, this 90\% to 94\% agreement between state-of-the-art agents and the reference outputs confirms the ground-truth correctness of the generated benchmarks while motivating format-tolerant grading.

Second, the remaining failures in S4 are genuine reasoning failures where the agent fails to look up or apply the right business rules, terminology, or tables (accounting for 61\% of Gemma's failures). Across all three models, these failures follow three patterns. First, the agent fails to retrieve or apply a secondary rule in $\mathcal{D}_{\text{doc}}$ during multi-hop search, such as filtering \texttt{qa-canary} accounts in Music store while missing a second document that excludes a six-hour outage window, or including negative credit balances in Insurance FIFO lockbox allocation when the policy memo restricts prior dues to positive balances. Second, the agent queries a partial or stale table over the authoritative table, such as summing a five-minute sensor table in Retail-1000 over \texttt{footfall\_daily}. Third, the agent applies incomplete date or status filters.

\smallskip
\noindent \textbf{Verifying answers and challenging state-of-the-art agents.}
\definecolor{failfg}{RGB}{180, 32, 32}
\newcommand{\fromlog}[1]{\textcolor{glossfg}{#1}}
\newcommand{\fromfail}[1]{\textcolor{failfg}{#1}}
Pushing state-of-the-art agents beyond moderate-complexity tasks requires harder multi-event questions, where annotators who see only the final database can no longer derive reliable reference answers. On a 1K-table warehouse, when candidate gold programs differ by a narrow margin, human annotators inspecting only the final database struggle to determine or defend the reference answer. In the Retail-1000 task below, the initial candidate gold program reads current loyalty tiers from \texttt{loyalty\_members} and makes the same mistake that Opus makes in all 12 of its runs.

\begin{tcolorbox}[colback=gray!4,colframe=gray!50,boxrule=0.4pt,left=3pt,right=3pt,top=2pt,bottom=2pt,fontupper=\footnotesize]
\textbf{Composite task (Retail-1000).} \emph{``\ldots{} as of August~21, 2026, provide an operational summary across all six market zones \ldots{} Total Store Count, Company Staff Count, Bronze Loyalty Members, \ldots{} and Customer Merge Rate.''}\\[2pt]
\circled{1} \textbf{\fromdoc{Complexities.}} Combines store status, company staffing, and acquired-customer deduplication (\texttt{bb\_tbl\_customer\_merge\_map}, \ldots{}) with \fromdoc{August~21 loyalty tiers}. An August~26 event overwrites \fromdoc{\texttt{loyalty\_members.tier\_id}} in place, while \fromdoc{\texttt{snap\_dim\_customer\_history}} retains \fromdoc{(\texttt{valid\_from}, \texttt{valid\_to})} intervals.\\
\circled{2} \textbf{\fromlog{Provenance audit ($\mathcal{L}_{\text{log}}$).}} The candidate gold program reads current \texttt{loyalty\_members.tier\_id}. Critics trace the \fromlog{logged August~26 re-tiering event} (where member~802 moves to Bronze) and replace it with a point-in-time join on \fromlog{\texttt{snap\_dim\_customer\_history}} (\fromlog{826} versus 880 Bronze members).\\
\circled{3} \textbf{\fromfail{Agent failure (Opus).}} \fromfail{Every run counts \texttt{loyalty\_members} by current \texttt{tier\_id}} and matches the candidate gold (880 Bronze members). Two runs see \texttt{snap\_dim\_customer\_history} in tool output, yet no run queries it.
\end{tcolorbox}

\system resolves such disputes because $\mathcal{L}_{\text{log}}$ records the causal provenance of every row and schema change. Critics in the multi-agent debate trace the discrepancy directly to the logged August~26 re-tiering event and the warehouse snapshot guide in $\mathcal{D}_{\text{doc}}$, and revise the candidate gold program to join the historical snapshot table at the target date. Conversely, when $\mathcal{L}_{\text{log}}$ shows that a question admits two valid interpretations under the logged rules (such as per-ticket versus aggregate rounding when backing out 7.25\% inclusive tax at acquired stores, shifting totals by \$0.24 to \$0.35), the panel adds explicit specifications to the question to eliminate the ambiguity.

Once $\mathcal{L}_{\text{log}}$ defends the reference answer, composite tasks reliably expose reasoning limits in state-of-the-art agents. On the task above, Opus computes the other seven columns correctly in all 12 runs, yet never checks whether tier assignments changed after the target date, even when the history snapshot appears in its tool output. DeepSeek also fails all 12 runs. On the 11 multi-rule composite tasks of Retail-1000, S3 accuracy falls from 91\% on Opus to 76\% on DeepSeek and 33\% on Gemma, showing how $\mathcal{L}_{\text{log}}$ enables harder verified benchmarks that drive data-agent improvement.

\section{The Boundaries of the Simulation}
\label{sec:realism}
\label{sec:eval_realism}

\begin{table}[t]
\centering\small
\caption{Data realism. The top panel compares Banking with the real Czech bank data of BIRD \texttt{financial}. The bottom panel tests five stylized facts across all data estates against real references. Shading follows \realismbar.}
\label{tab:realism}
\setlength{\tabcolsep}{5pt}
\renewcommand{\arraystretch}{1.12}
\begin{tabular}{@{}>{\raggedright\arraybackslash}p{\dimexpr\textwidth-5.6cm-5\tabcolsep\relax}>{\raggedleft\arraybackslash}p{2.4cm}>{\raggedleft\arraybackslash}p{3.2cm}}
\toprule
\textbf{Banking vs.\ BIRD \texttt{financial}} & \textbf{BIRD} & \textbf{\system} \\
\midrule
Weekend share of transactions & 0.291 & \qcell{qhigh}{0.283} \\
Share of round-hundred amounts & 0.247 & \qcell{qhigh!85!qmid}{0.227} \\
Transactions per account per 30 days (median) & 5.4 & \qcell{qhigh!65!qmid}{6.6} \\
Top-20\% accounts' share of transactions & 0.37 & \qcell{qhigh!60!qmid}{0.29} \\
Amount tail, p99/p50 & 21.6 & \qcell{qhigh!50!qmid}{15.8} \\
Transaction code mix (\texttt{type}/\texttt{operation}/\texttt{k\_symbol}), TVD & -- & \qcell{fiderr}{0.44 / 0.45 / 0.37} \\
Account-days whose balances reconcile & 1.000 & \qcell{fiderr!80!qmid}{0.923} \\
\bottomrule
\end{tabular}

\medskip
\setlength{\tabcolsep}{4.5pt}
\resizebox{\textwidth}{!}{%
\begin{tabular}{@{}lrrrrrrrr}
\toprule
\textbf{Stylized fact} & \textbf{Real} & \textbf{Retail-10} & \textbf{Retail-100} & \textbf{Retail-1000} & \textbf{Music store} & \textbf{Banking} & \textbf{Logistics} & \textbf{Insurance} \\
\midrule
Weekend/weekday daily volume & 0.75\textsuperscript{O} / 1.03\textsuperscript{B} & \qcell{fiderr!75!qmid}{0.99} & \qcell{fiderr!80!qmid}{1.02} & \qcell{fiderr!80!qmid}{1.01} & \qcell{qhigh!65!qmid}{1.11} & \qcell{qhigh}{0.99} & \qcell{fiderr}{1.17} & \qcell{qhigh!75!qmid}{1.04} \\
Top-20\% customers' share of activity & 0.73\textsuperscript{K} / 0.37\textsuperscript{B} & \qcell{fiderr!85!qmid}{0.29} & \qcell{fiderr!80!qmid}{0.30} & \qcell{fiderr!70!qmid}{0.34} & \qcell{qmid}{0.40} & \qcell{qhigh!60!qmid}{0.29} & \qcell{fiderr}{0.23} & \qcell{qhigh}{0.37} \\
Human price endings & 0.85\textsuperscript{O} / 1.00\textsuperscript{C} & \qcell{qhigh}{1.00} & \qcell{qhigh}{1.00} & \qcell{qhigh}{1.00} & \qcell{qhigh}{1.00} & \qcell{qhigh!85!qmid}{0.23} & \qcell{qhigh!65!qmid}{0.63} & \qcell{qhigh!50!qmid}{0.50} \\
Accounting identities hold & 1.00\textsuperscript{B} & \qcell{qhigh}{1.00} & \qcell{qhigh}{1.00} & \qcell{qhigh}{1.00} & \qcell{qhigh}{1.00} & \qcell{fiderr!80!qmid}{0.92} & \qcell{qhigh}{1.00} & \qcell{qhigh}{1.00} \\
Amount tail, p99/p50 & 10.0\textsuperscript{O} / 21.6\textsuperscript{B} & \qcell{qhigh!40!qmid}{3.5} & \qcell{qhigh!35!qmid}{3.2} & \qcell{qmid}{2.7} & \qcell{qhigh!80!qmid}{7.5} & \qcell{qhigh!70!qmid}{15.8} & \qcell{qhigh!90!qmid}{16.9} & \qcell{fiderr!80!qmid}{6.3} \\
\bottomrule
\end{tabular}%
}
\\[3pt]
\parbox{\textwidth}{\footnotesize \textit{Notes.} TVD denotes total variation distance. Superscripts mark real references, with B for BIRD \texttt{financial} (0.25 round-hundred share on Banking price endings), O for Olist orders, K for consumer-goods panels, and C for Chinook.}
\end{table}

Pipeline correctness evaluates whether each data estate satisfies its scenario specification. We can also ask whether the data values \system generates are realistic. However, since real-world enterprises are open-ended, in practice every simulation has a {\em boundary} that delineates what is in the simulation scope and what is not.

We built \system to evaluate data agents on enterprise tasks that involve data integration and deep data research, such as discovering authoritative tables, resolving business terms, and writing SQL over a complex warehouse. As such, our goal is to develop data estates whose complexity is largely at the metadata level.  \emph{The actual data values in the databases are much less important.}

For data integration tasks, row-level value distributions matter very little as long as foreign keys join cleanly and column profiles match the catalog descriptions. Deep data research tasks (such as testing whether weekend discounts increased cafe visits or how weather affects daily revenue) depend on statistical relationships in the data.
However,
  \system events encode those discount or weather effects directly in the simulation distributions, which plants the target relationship in the data for the agent to uncover. Distributional realism thus remains secondary for both task families. This would change if we expand the simulation boundary of \system to consider simulations that drive operational business decisions, such as optimizing marketing campaigns across competing business objectives, which we leave to future work.

Still, because a simulated enterprise has no row-level ground truth, we evaluate data realism with deterministic, scale-free distributional checks in two ways. Table~\ref{tab:realism} summarizes the results, and Appendix~\ref{app:realism} and Table~\ref{tab:realism_full} define the metrics and report the complete evaluation.
First, because Banking reuses the schema of BIRD \texttt{financial}~\citep{li2023can} (1.06M real Czech bank transactions that \system never accesses), we compare simulated and real tables column by column. Two factors explain the divergences. First, an early event generates a 2026-style codebook that books cash and fee withdrawals under dedicated types and treats card payments as routine, unlike the 1990s Czech bank where card withdrawals are rare. Second, the monthly payroll event computes salary-credit balances from the account table, which lags behind the transaction ledger when payroll runs, leaving affected account-days with a stale starting balance and lowering the Banking accounting-identity score in the bottom panel.

Second, following standard simulation validation~\citep{windrum2007empirical}, we test five stylized facts across all seven data estates against four real reference datasets, namely BIRD \texttt{financial}~\citep{li2023can}, Olist e-commerce orders~\citep{olist2018}, the Chinook media catalog, and consumer-goods purchase panels~\citep{kim2017pareto}. Deterministic scripts compute all metrics over SQLite, where \emph{top-20\% share} is the activity share of the top fifth of customers and \emph{amount tail} is the p99/p50 ratio of absolute amounts. \system reproduces facts stated in event programs and schema rules. LLM-generated price lists follow human price endings, and SQL line-item rollups preserve accounting identities by construction. Weekly seasonality and customer concentration~\citep{kim2017pareto} appear when requested in the scenario prompt, whereas unprompted scenarios draw daily counts from fixed ranges and sample accounts uniformly, and 16-item cafe menus bound Retail receipt tails.

\section{Conclusion and Future Work}
\label{sec:conclusion}

We presented \system, a platform that creates customizable data estates and benchmarks for data agents by simulating how business events produce and modify corporate data over time. By compiling reusable event types into programs over simulation primitives and scheduling routine and disruption events on a shared virtual calendar, \system keeps relational tables, policy documents, and catalog entries mutually consistent while recording the complete provenance of the data estate in a simulation log. Our evaluation across seven data estates and 256 tasks in five domains shows that \system reliably materializes data estates up to 1K operational tables, derives accurate gold programs from the simulation log without manual annotation, and produces benchmarks that differentiate both data-context settings and model tiers.

The current benchmark generator represents a first step. Although many of the generated tasks are relatively simple for frontier models, the benchmarks already contain multi-event composite tasks that cause state-of-the-art agents to fail while the simulation log verifies the reference answers. We will continue iterating on the benchmark generator to produce more challenging tasks by composing longer chains of business events, deeper multi-document rules, and point-in-time state changes, and by using agent failure patterns to steer task synthesis. On the simulation side, we plan to add compiler guardrails against cross-column and cross-row dependency violations and expand our complexity library across new enterprise domains. Looking ahead, \system enables benchmarks that co-evolve with data agents, simulated data estates that provide verifiable rewards for agent training, and simulations where LLM agents act as employees and drive business events.

\newpage
\bibliography{reference}

@article{fu2025autonomous,
  title={Autonomous Data Agents: A New Opportunity for Smart Data},
  author={Fu, Yanjie and Wang, Dongjie and Ying, Wangyang and Wang, Xinyuan and Zhang, Xiangliang and Liu, Huan and Pei, Jian},
  journal={arXiv preprint arXiv:2509.18710},
  year={2025}
}

@article{zhu2025survey,
  title={A Survey of Data Agents: Emerging Paradigm or Overstated Hype?},
  author={Zhu, Yizhang and Wang, Liangwei and Yang, Chenyu and Lin, Xiaotian and Li, Boyan and Zhou, Wei and Liu, Xinyu and Peng, Zhangyang and Luo, Tianqi and Li, Yu and others},
  journal={arXiv preprint arXiv:2510.23587},
  year={2025}
}

@article{allen1983maintaining,
  title={Maintaining Knowledge about Temporal Intervals},
  author={Allen, James F.},
  journal={Communications of the {ACM}},
  volume={26},
  number={11},
  pages={832--843},
  year={1983}
}

@article{egg2025dabstep,
  title={{DABstep}: Data Agent Benchmark for Multi-step Reasoning},
  author={Egg, Alex and Goyanes, Martin Iglesias and Kingma, Friso and Mora, Andreu and von Werra, Leandro and Wolf, Thomas},
  journal={arXiv preprint arXiv:2506.23719},
  year={2025}
}

@inproceedings{lai2026kramabench,
  title={{KramaBench}: A Benchmark for {AI} Systems on Data-to-Insight Pipelines over Data Lakes},
  author={Lai, Eugenie and Vitagliano, Gerardo and Zhang, Ziyu and Chabra, Om and Sudhir, Sivaprasad and Zeng, Anna and Zabreyko, Anton and Li, Chenning and Kossmann, Ferdi and Ding, Jialin and others},
  booktitle={International Conference on Learning Representations (ICLR)},
  year={2026}
}

@inproceedings{lei2026dacomp,
  title={{DAComp}: Benchmarking Data Agents across the Full Data Intelligence Lifecycle},
  author={Lei, Fangyu and Meng, Jinxiang and Huang, Yiming and Zhang, Yitong and Luo, Jianwen and Zou, Xin and Yang, Ruiyi and Shi, Wenbo and Gao, Yan and He, Shizhu and others},
  booktitle={International Conference on Learning Representations (ICLR)},
  year={2026}
}

@inproceedings{lei2025spider,
  title={Spider 2.0: Evaluating Language Models on Real-World Enterprise Text-to-{SQL} Workflows},
  author={Lei, Fangyu and Chen, Jixuan and Ye, Yuxiao and Cao, Ruisheng and Shin, Dongchan and Su, Hongjin and Suo, Zhaoqing and Gao, Hongcheng and Hu, Wenjing and Yin, Pengcheng and others},
  booktitle={International Conference on Learning Representations (ICLR)},
  year={2025}
}

@inproceedings{li2023can,
  title={Can {LLM} Already Serve as a Database Interface? {A} {BIg} Bench for Large-Scale Database Grounded Text-to-{SQLs}},
  author={Li, Jinyang and Hui, Binyuan and Qu, Ge and Yang, Jiaxi and Li, Binhua and Li, Bowen and Wang, Bailin and Qin, Bowen and Geng, Ruiying and Huo, Nan and others},
  booktitle={Advances in Neural Information Processing Systems (NeurIPS)},
  volume={36},
  pages={42330--42357},
  year={2023}
}

@article{wang2026lakeqa,
  title={{LakeQA}: An Exploratory {QA} Benchmark over a Million-Scale Data Lake},
  author={Wang, Haonan and Liu, Jiaxiang and Liu, Yurong and Wijaya, Austin Senna and Zhou, Tianle and Wu, Eden and Chen, Yijia and You, Wanting and Vir, Reya and Pinto, Daniela and others},
  journal={arXiv preprint arXiv:2606.10460},
  year={2026}
}

@article{zhang2026coda,
  title={{CODA-BENCH}: Can Code Agents Handle Data-Intensive Tasks?},
  author={Zhang, Yuxin and Fan, Ju and Fan, Meihao and Zhang, Shaolei and Du, Xiaoyong},
  journal={arXiv preprint arXiv:2606.15300},
  year={2026}
}

@inproceedings{northcutt2021pervasive,
  title={Pervasive Label Errors in Test Sets Destabilize Machine Learning Benchmarks},
  author={Northcutt, Curtis G. and Athalye, Anish and Mueller, Jonas},
  booktitle={Proceedings of the Neural Information Processing Systems Track on Datasets and Benchmarks},
  year={2021}
}

@inproceedings{weber2024varierr,
  title={{VariErr NLI}: Separating Annotation Error from Human Label Variation},
  author={Weber-Genzel, Leon and Peng, Siyao and De Marneffe, Marie-Catherine and Plank, Barbara},
  booktitle={Proceedings of the 62nd Annual Meeting of the Association for Computational Linguistics (ACL)},
  pages={2256--2269},
  year={2024}
}

@inproceedings{long2024llms,
  title={On {LLMs}-Driven Synthetic Data Generation, Curation, and Evaluation: A Survey},
  author={Long, Lin and Wang, Rui and Xiao, Ruixuan and Zhao, Junbo and Ding, Xiao and Chen, Gang and Wang, Haobo},
  booktitle={Findings of the Association for Computational Linguistics: ACL 2024},
  pages={11065--11082},
  year={2024}
}

@inproceedings{park2023generative,
  title={Generative Agents: Interactive Simulacra of Human Behavior},
  author={Park, Joon Sung and O'Brien, Joseph and Cai, Carrie Jun and Morris, Meredith Ringel and Liang, Percy and Bernstein, Michael S.},
  booktitle={Proceedings of the 36th Annual {ACM} Symposium on User Interface Software and Technology (UIST)},
  pages={1--22},
  year={2023}
}

@inproceedings{ni2026survey,
  title={A Survey on {LLM}-based Conversational User Simulation},
  author={Ni, Bo and Wang, Yu and Wang, Leyao and Kveton, Branislav and Dernoncourt, Franck and Xia, Yu and Chen, Hongjie and Luera, Reuben and Basu, Samyadeep and Mukherjee, Subhojyoti and others},
  booktitle={Proceedings of the 19th Conference of the European Chapter of the Association for Computational Linguistics (EACL)},
  pages={4266--4301},
  year={2026}
}

@article{lao2025sembench,
  title={{SemBench}: A Benchmark for Semantic Query Processing Engines},
  author={Lao, Jiale and Zimmerer, Andreas and Ovcharenko, Olga and Cong, Tianji and Russo, Matthew and Vitagliano, Gerardo and Cochez, Michael and {\"O}zcan, Fatma and Gupta, Gautam and Hottelier, Thibaud and others},
  journal={Proceedings of the VLDB Endowment},
  volume={19},
  number={8},
  pages={1754--1767},
  year={2026}
}

@inproceedings{jing2025dsbench,
  title={{DSBench}: How Far Are Data Science Agents from Becoming Data Science Experts?},
  author={Jing, Liqiang and Huang, Zhehui and Wang, Xiaoyang and Yao, Wenlin and Yu, Wenhao and Ma, Kaixin and Zhang, Hongming and Du, Xinya and Yu, Dong},
  booktitle={International Conference on Learning Representations},
  volume={2025},
  pages={32597--32649},
  year={2025}
}

@inproceedings{gray1994quickly,
  title={Quickly Generating Billion-Record Synthetic Databases},
  author={Gray, Jim and Sundaresan, Prakash and Englert, Susanne and Baclawski, Ken and Weinberger, Peter J.},
  booktitle={Proceedings of the 1994 {ACM} {SIGMOD} International Conference on Management of Data},
  pages={243--252},
  year={1994}
}

@inproceedings{touchstone18,
  author={Li, Yuming and Zhang, Rong and Yang, Xiaoyan and Zhang, Zhenjie and Zhou, Aoying},
  title={Touchstone: Generating Enormous {Query-Aware} Test Databases},
  booktitle={2018 {USENIX} Annual Technical Conference ({USENIX} {ATC} 18)},
  pages={575--586},
  year={2018}
}

@article{li2025omnisql,
author = {Li, Haoyang and Wu, Shang and Zhang, Xiaokang and Huang, Xinmei and Zhang, Jing and Jiang, Fuxin and Wang, Shuai and Zhang, Tieying and Chen, Jianjun and Shi, Rui and Chen, Hong and Li, Cuiping},
title = {{OmniSQL}: Synthesizing High-Quality Text-to-{SQL} Data at Scale},
year = {2025},
publisher = {VLDB Endowment},
volume = {18},
number = {11},
issn = {2150-8097},
url = {https://doi.org/10.14778/3749646.3749723},
doi = {10.14778/3749646.3749723},
journal = {Proc. VLDB Endow.},
pages = {4695–4709},
numpages = {15}
}

@inproceedings{xu2019modeling,
  title={Modeling Tabular Data Using Conditional {GAN}},
  author={Xu, Lei and Skoularidou, Maria and Cuesta-Infante, Alfredo and Veeramachaneni, Kalyan},
  booktitle={Advances in Neural Information Processing Systems (NeurIPS)},
  volume={32},
  year={2019}
}

@inproceedings{zhang2024mixed,
  title={Mixed-Type Tabular Data Synthesis with Score-Based Diffusion in Latent Space},
  author={Zhang, Hengrui and Zhang, Jiani and Shen, Zhengyuan and Srinivasan, Balasubramaniam and Qin, Xiao and Faloutsos, Christos and Rangwala, Huzefa and Karypis, George},
  booktitle={International Conference on Learning Representations (ICLR)},
  year={2024}
}

@article{jin2025elt,
  title={{ELT-Bench}: An End-to-End Benchmark for Evaluating {AI} Agents on {ELT} Pipelines},
  author={Jin, Tengjun and Zhu, Yuxuan and Kang, Daniel},
  journal={Proceedings of the VLDB Endowment},
  volume={19},
  number={2},
  pages={84--98},
  year={2025}
}

@article{dechter1991temporal,
  title={Temporal Constraint Networks},
  author={Dechter, Rina and Meiri, Itay and Pearl, Judea},
  journal={Artificial Intelligence},
  volume={49},
  number={1--3},
  pages={61--95},
  year={1991}
}

@article{bodensohn2025unveiling,
  title={Unveiling Challenges for {LLMs} in Enterprise Data Engineering},
  author={Bodensohn, Jan-Micha and Brackmann, Ulf and Vogel, Liane and Sanghi, Anupam and Binnig, Carsten},
  journal={Proceedings of the VLDB Endowment},
  volume={19},
  number={2},
  pages={196--209},
  year={2025},
  publisher={VLDB Endowment}
}

@article{jin2026pervasive,
  title={Pervasive Annotation Errors Break Text-to-{SQL} Benchmarks and Leaderboards},
  author={Jin, Tengjun and Choi, Yoojin and Zhu, Yuxuan and Kang, Daniel},
  journal={arXiv preprint arXiv:2601.08778},
  year={2026}
}

@article{zhang2026agentic,
  title={An Agentic Approach to Metadata Reasoning},
  author={Zhang, Jiani and Arik, Sercan O and Arad, Cosmin and Ozcan, Fatma and Halevy, Alon},
  journal={arXiv preprint arXiv:2604.20144},
  year={2026}
}

@misc{olist2018,
  author={Olist and Sionek, Andr{\'e}},
  title={Brazilian E-Commerce Public Dataset by {Olist}},
  year={2018},
  publisher={Kaggle},
  howpublished={\url{https://www.kaggle.com/datasets/olistbr/brazilian-ecommerce}},
  doi={10.34740/KAGGLE/DSV/195341}
}

@article{kim2017pareto,
  title={The {Pareto} Rule for Frequently Purchased Packaged Goods: An Empirical Generalization},
  author={Kim, Baek Jung and Singh, Vishal and Winer, Russell S},
  journal={Marketing Letters},
  volume={28},
  number={4},
  pages={491--507},
  year={2017},
  publisher={Springer}
}

@article{windrum2007empirical,
  title={Empirical Validation of Agent-Based Models: Alternatives and Prospects},
  author={Windrum, Paul and Fagiolo, Giorgio and Moneta, Alessio},
  journal={Journal of Artificial Societies and Social Simulation},
  volume={10},
  number={2},
  pages={8},
  year={2007}
}

@misc{halevy2026toaster,
  title={The Toaster Problem: {A} North Star Challenge for Data Agents},
  author={Halevy, Alon},
  howpublished={\url{https://alonhalevy.substack.com/p/the-toaster-problem}},
  year={2026}
}

@article{flynt2026orgforge,
  title   = {{OrgForge}: A Multi-Agent Simulation Framework for Verifiable Synthetic Corporate Corpora},
  author  = {Flynt, Jeffrey},
  journal = {arXiv preprint arXiv:2603.14997},
  year    = {2026},
  url     = {https://arxiv.org/abs/2603.14997}
}

@article{huang2025crmarenapro,
  title={{CRMArena-Pro}: Holistic Assessment of {LLM} Agents Across Diverse Business Scenarios and Interactions},
  author={Huang, Kung-Hsiang and Prabhakar, Akshara and Thorat, Onkar and Agarwal, Divyansh and Choubey, Prafulla Kumar and Mao, Yixin and Savarese, Silvio and Xiong, Caiming and Wu, Chien-Sheng},
  journal={arXiv preprint arXiv:2505.18878},
  year={2025}
}

@article{xu2025theagentcompany,
  title={Theagentcompany: benchmarking llm agents on consequential real world tasks},
  author={Xu, Frank Fangzheng and Song, Yufan and Li, Boxuan and Tang, Yuxuan and Jain, Kritanjali and Bao, Mengxue and Wang, Zora and Zhou, Xuhui and Guo, Zhitong and Cao, Murong and others},
  journal={Advances in Neural Information Processing Systems},
  volume={38},
  year={2026}
}

@article{gruenbaum2026era,
  title   = {The {Era by Eon} Benchmark: A Generated Enterprise Estate with Exact Ground Truth for Benchmarking {LLM} Agents},
  author  = {Gruenbaum, Benjamin and Porat, Doron and Natanzon, Assaf and Zavida, Roy and Dinachi, Chen and Itzahary, Or},
  journal = {arXiv preprint arXiv:2609.09853},
  year    = {2026},
  url     = {https://arxiv.org/abs/2609.09853}
}

@inproceedings{burattin2016plg2,
  title     = {{PLG2}: Multiperspective Process Randomization with Online and Offline Simulations},
  author    = {Burattin, Andrea},
  booktitle = {Proceedings of the BPM Demo Track (CEUR Workshop Proceedings, Vol. 1789)},
  pages     = {1--6},
  year      = {2016},
  url       = {https://arxiv.org/abs/1506.08415}
}

@inproceedings{kirchdorfer2024agentsimulator,
  title     = {{AgentSimulator}: An Agent-based Approach for Data-driven Business Process Simulation},
  author    = {Kirchdorfer, Lukas and Bl{\"u}mel, Robert and Kampik, Timotheus and van der Aa, Han and Stuckenschmidt, Heiner},
  booktitle = {Proceedings of the 6th International Conference on Process Mining (ICPM)},
  pages     = {97--104},
  year      = {2024},
  doi       = {10.1109/ICPM63005.2024.10680660},
  url       = {https://arxiv.org/abs/2408.08571}
}

@article{camargo2020simod,
  title   = {Automated Discovery of Business Process Simulation Models from Event Logs},
  author  = {Camargo, Manuel and Dumas, Marlon and Gonz{\'a}lez-Rojas, Oscar},
  journal = {Decision Support Systems},
  volume  = {134},
  pages   = {113284},
  year    = {2020},
  doi     = {10.1016/j.dss.2020.113284},
  url     = {https://arxiv.org/abs/1910.05404}
}

@article{sanghi2018hydra,
  title   = {{HYDRA}: A Dynamic Big Data Regenerator},
  author  = {Sanghi, Anupam and Sood, Raghav and Singh, Dharmendra and Haritsa, Jayant R. and Tirthapura, Srikanta},
  journal = {Proceedings of the VLDB Endowment},
  volume  = {11},
  number  = {12},
  pages   = {1974--1977},
  year    = {2018},
  doi     = {10.14778/3229863.3236238}
}

@inproceedings{patki2016sdv,
  title     = {The Synthetic Data Vault},
  author    = {Patki, Neha and Wedge, Roy and Veeramachaneni, Kalyan},
  booktitle = {IEEE International Conference on Data Science and Advanced Analytics (DSAA)},
  pages     = {399--410},
  year      = {2016},
  doi       = {10.1109/DSAA.2016.49}
}

@inproceedings{pang2024clavaddpm,
author = {Pang, Wei and Shafieinejad, Masoumeh and Liu, Lucy and Hazlewood, Stephanie and He, Xi},
title = {ClavaDDPM: multi-relational data synthesis with cluster-guided diffusion models},
year = {2024},
booktitle = {Proceedings of the 38th International Conference on Neural Information Processing Systems},
articleno = {2657},
numpages = {27}
}

@inproceedings{yang2024sense,
    title = "Synthesizing Text-to-{SQL} Data from Weak and Strong {LLM}s",
    author = "Yang, Jiaxi  and
      Hui, Binyuan  and
      Yang, Min  and
      Yang, Jian  and
      Lin, Junyang  and
      Zhou, Chang",
    booktitle = "Proceedings of the 62nd Annual Meeting of the {A}ssociation for {C}omputational {L}inguistics (Volume 1: Long Papers)",
    year = "2024",
    url = "https://aclanthology.org/2024.acl-long.425/",
    doi = "10.18653/v1/2024.acl-long.425",
    pages = "7864--7875",
}

@article{habibollah2026synsql,
  title   = {{SynSQL}: Synthesizing Relational Databases for Robust Evaluation of Text-to-{SQL} Systems},
  author  = {Habibollah, Mohammadamin and Rafiei, Davood},
  journal = {arXiv preprint arXiv:2604.27261},
  year    = {2026},
  url     = {https://arxiv.org/abs/2604.27261}
}

@article{dai2025lakegen,
  title   = {{LakeGen}: A {LLM}-based Tabular Corpus Generator for Evaluating Dataset Discovery in Data Lakes},
  author  = {Dai, Zhenwei and Lei, Chuan and Katsifodimos, Asterios and Qin, Xiao and Faloutsos, Christos and Rangwala, Huzefa},
  journal = {arXiv preprint arXiv:2507.04687},
  year    = {2025},
  url     = {https://arxiv.org/abs/2507.04687}
}

@article{arocena2015bart,
  title   = {Messing Up with {BART}: Error Generation for Evaluating Data-Cleaning Algorithms},
  author  = {Arocena, Patricia C. and Glavic, Boris and Mecca, Giansalvatore and Miller, Ren{\'e}e J. and Papotti, Paolo and Santoro, Donatello},
  journal = {Proceedings of the VLDB Endowment},
  volume  = {9},
  number  = {2},
  pages   = {36--47},
  year    = {2015},
  url     = {http://www.vldb.org/pvldb/vol9/p36-arocena.pdf}
}

@article{zhang2025evoschema,
  title   = {{EvoSchema}: Towards Text-to-{SQL} Robustness Against Schema Evolution},
  author  = {Zhang, Tianshu and Qian, Kun and Sahai, Siddhartha and Tian, Yuan and Garg, Shaddy and Sun, Huan and Li, Yunyao},
  journal = {Proceedings of the VLDB Endowment},
  volume  = {18},
  number  = {10},
  pages   = {3655--3668},
  year    = {2025},
  doi     = {10.14778/3748191.3748222},
  url     = {https://arxiv.org/abs/2603.10697}
}

@InProceedings{kambhampati2024llmmodulo,
  title = 	 {Position: {LLM}s Can’t Plan, But Can Help Planning in {LLM}-Modulo Frameworks},
  author =       {Kambhampati, Subbarao and Valmeekam, Karthik and Guan, Lin and Verma, Mudit and Stechly, Kaya and Bhambri, Siddhant and Saldyt, Lucas Paul and B Murthy, Anil},
  booktitle = 	 {Proceedings of the 41st International Conference on Machine Learning},
  pages = 	 {22895--22907},
  year = 	 {2024},
  volume = 	 {235},
  url = 	 {https://proceedings.mlr.press/v235/kambhampati24a.html},
}

@article{chen2024beaver,
  title   = {{BEAVER}: An Enterprise Benchmark for Text-to-{SQL}},
  author  = {Chen, Peter Baile and Wenz, Fabian and Zhang, Yi and Yang, Devin and Choi, Justin and Tatbul, Nesime and Cafarella, Michael and Demiralp, {\c{C}}a{\u{g}}atay and Stonebraker, Michael},
  journal = {arXiv preprint arXiv:2409.02038},
  year    = {2024},
  url     = {https://arxiv.org/abs/2409.02038}
}

@inproceedings{hao2023rap,
    title = "Reasoning with Language Model is Planning with World Model",
    author = "Hao, Shibo  and
      Gu, Yi  and
      Ma, Haodi  and
      Hong, Joshua  and
      Wang, Zhen  and
      Wang, Daisy  and
      Hu, Zhiting",
    booktitle = "Proceedings of the 2023 Conference on Empirical Methods in Natural Language Processing",
    year = "2023",
    url = "https://aclanthology.org/2023.emnlp-main.507/",
    doi = "10.18653/v1/2023.emnlp-main.507",
    pages = "8154--8173"
}

@article{tang2024worldcoder,
  title={Worldcoder, a model-based llm agent: Building world models by writing code and interacting with the environment},
  author={Tang, Hao and Key, Darren and Ellis, Kevin},
  journal={Advances in Neural Information Processing Systems},
  volume={37},
  pages={70148--70212},
  year={2024}
}

@article{zeng2026minicorp,
  title   = {{MiniCorp}: The Last Mile to an {AI} Agent Corporation},
  author  = {Zeng, Jingying and Dai, Zhenwei and Li, Jinning and Shin, Changho and Zhang, Dylan and Lu, Yuxuan and He, Qi and Wang, Dakuo and Chang, Kai-Wei},
  journal = {arXiv preprint arXiv:2610.05912},
  year    = {2026},
  url     = {https://arxiv.org/abs/2610.05912}
}

@inproceedings{ma2026can,
  title={Can {AI} Agents Answer Your Data Questions? A Benchmark for Data Agents},
  author={Ma, Ruiying and Shankar, Shreya and Chen, Ruiqi and Chintamaneni, Adit and Lin, Yiming and Zeighami, Sepanta and Ghosh, Rajoshi and Gupta, Abhinav and Gupta, Anushrut and Gopal, Tanmai and others},
  booktitle={Proceedings of the 2026 Conference on Empirical Methods in Natural Language Processing},
  pages={42651--42677},
  year={2026}
}

@article{abdelaal2023rein,
  title={{REIN}: A Comprehensive Benchmark Framework for Data Cleaning Methods in {ML} Pipelines},
  author={Abdelaal, Mohamed and Hammacher, Christian and Schoening, Harald},
  journal={arXiv preprint arXiv:2302.04702},
  year={2023}
}

@article{chang2023dr,
  title={{Dr. Spider}: A Diagnostic Evaluation Benchmark Towards Text-to-{SQL} Robustness},
  author={Chang, Shuaichen and Wang, Jun and Dong, Mingwen and Pan, Lin and Zhu, Henghui and Li, Alexander Hanbo and Lan, Wuwei and Zhang, Sheng and Jiang, Jiarong and Lilien, Joseph and others},
  journal={arXiv preprint arXiv:2301.08881},
  year={2023}
}

@article{luoma2025snails,
  title={{SNAILS}: Schema Naming Assessments for Improved {LLM}-Based {SQL} Inference},
  author={Luoma, Kyle and Kumar, Arun},
  journal={Proceedings of the ACM on Management of Data},
  volume={3},
  number={1},
  pages={1--26},
  year={2025},
  publisher={ACM, New York, NY, USA}
}

@article{berka2000guide,
  author = {Berka, Petr and Sochorov{'a}, Marta},
  title = {Guide to the Financial Data Set},
  journal = {The Discovery Challenge at PKDD},
  year = {2000}
}

\newpage
\appendix

\section{Scenario Descriptions and Simulation Instructions}
\label{app:scenarios}

This appendix summarizes the seven data estates across the five domains of Section~\ref{sec:eval_scenario} and Table~\ref{tab:estates}.
\begin{itemize}[leftmargin=.5cm,itemsep=2pt]
\item \textbf{Retail at three scales.} Retail-10, Retail-100, and Retail-1000 model a specialty coffee business across three operational scales.
\begin{itemize}[leftmargin=.4cm,itemsep=1pt,topsep=1pt]
\item \textbf{Retail-10} covers counter sales, register reconciliation, loyalty discounts, staff shifts, suppliers, and a daily sales summary.
\item \textbf{Retail-100} keeps the single store and its guest traffic while adding 14 back-office subsystems and recurring departmental reports, including multi-tender POS, online and catering orders, recipe-driven stock depletion over 40 ingredients, three-way-matched procurement, roasting, maintenance, HACCP logs, payroll, and a 30-account general ledger.
\item \textbf{Retail-1000} scales to a regional chain with eight company-owned stores, 12 franchised stores, a central roastery, an e-commerce storefront, a distribution center, and an acquired six-store competitor whose tax-inclusive POS and weekly payroll tables remain unintegrated. Vendor landing zones, a multi-layer analytics warehouse, and surrounding shards, exports, and analyst sandboxes account for the remaining tables.
\end{itemize}
\item \textbf{Music store.} Chinook Digital Media sells digital music tracks on the public Chinook schema (\url{https://github.com/lerocha/chinook-database}). Policy documents state three implicit business rules. Revenue excludes nightly checkouts by QA canary accounts, the store anonymizes closed accounts in place while keeping their rows in the table, and the store issues off-database refunds for orders placed during a flash-sale outage.
\item \textbf{Banking.} Bohemia Commercial Bank, built on the Czech bank schema~\citep{berka2000guide} in BIRD \texttt{financial}~\citep{li2023can}, serves 200 clients across 12 districts. Transaction types, payment symbols, statement frequencies, and employment statuses use Czech codes defined in the bank codebook. The codebook also defines two KPIs and the statement-fee schedule. On Day~45, a steelworks closure leaves about 40\% of one district's employed clients unemployed and pushes their loans into arrears.
\item \textbf{Logistics.} The scenario adapts the ERP payment-to-invoice reconciliation task of \citet{bodensohn2025unveiling}. GlobalLogistics AG bills about 40 shipments per day, ingests statements from three house banks, and matches incoming payments to invoices. An auditor freezes a snapshot of the invoice table on Day~12, and a nightly reporting replica of bank transactions stops refreshing on Day~18. Treasury posts daily cash positions and monthly regional targets at coarser grains than invoices.
\item \textbf{Insurance.} Cascadia Mutual insures about 3,000 customers under auto and homeowners policies. Biweekly QA audits insert canary claims into the production database, lockbox remittances settle multiple invoices in a single lump sum, and a Day~48 telematics rating migration adds two unbackfilled columns to the insured-asset table for post-migration vehicles.
\end{itemize}

\subsection{Data Estate Composition}
\label{app:composition}

\begin{table}[!htbp]
\centering\small
\caption{Breakdown of tables, rows, and documents across the seven \system data estates, extending Table~\ref{tab:estates}.}
\label{tab:estate_composition}
\setlength{\tabcolsep}{0pt}
\renewcommand{\arraystretch}{1.08}
\begin{tabular*}{\textwidth}{@{\extracolsep{\fill}}l rrrrrrr@{}}
\toprule
 & \textbf{Retail-10} & \textbf{Retail-100} & \textbf{Retail-1000} & \textbf{Music store} & \textbf{Banking} & \textbf{Logistics} & \textbf{Insurance} \\
\midrule
Total tables & 33 & 272 & 1.4K & 11 & 25 & 60 & 126 \\
\quad report snapshots / views & 14/8 & 105/66 & 347/41 & 0/0 & 12/4 & 34/4 & 102/6 \\
\quad base tables & 11 & 101 & 1.0K & 11 & 9 & 22 & 18 \\
\qquad reference & 5 & 35 & 264 & 9 & 7 & 4 & 8 \\
\qquad transaction & 4 & 46 & 389 & 2 & 1 & 10 & 8 \\
\qquad log & 0 & 5 & 76 & 0 & 0 & 3 & 0 \\
\qquad aggregate & 2 & 15 & 247 & 0 & 1 & 5 & 2 \\
\qquad metadata & 0 & 0 & 28 & 0 & 0 & 0 & 0 \\
\midrule
Rows in transaction tables & 99\% & 95\% & 91\% & 67\% & 95\% & 96\% & 33\% \\
\midrule
Documents & 11 & 104 & 1.4K & 5 & 7 & 26 & 18 \\
\quad policy \& reference & 3 & 13 & 89 & 3 & 1 & 3 & 12 \\
\quad data documentation & 0 & 2 & 70 & 0 & 0 & 17 & 1 \\
\quad recurring reports & 4 & 73 & 1.2K & 0 & 4 & 4 & 4 \\
\quad event memos & 3 & 15 & 29 & 2 & 2 & 2 & 0 \\
\quad transcripts & 1 & 1 & 3 & 0 & 0 & 0 & 1 \\
\bottomrule
\end{tabular*}
\end{table}

Table~\ref{tab:estate_composition} breaks down the tables and documents of each data estate. We separate \emph{report snapshots}, which a \texttt{Query} operator materializes once per event instance (such as a weekly Insurance renewal audit), from \emph{base tables}, which \texttt{CreateSchema} or \texttt{AlterSchema} creates. We then group base tables into five roles by row grain.
\begin{itemize}[leftmargin=.5cm,itemsep=1pt]
\item \textbf{Reference.} One row per business entity, catalog item, contract, or code, such as customers, menu items, accounts, policies, and \texttt{dim\_*} tables.
\item \textbf{Transaction.} One row per business transaction or line item, such as receipts, invoice lines, payments, shifts, claims, and stock movements, including staged and snapshot copies.
\item \textbf{Log.} Machine- or system-generated append-only records, such as sensor readings, device heartbeats, web sessions, webhook and tracking events, audit trails, and pipeline runs.
\item \textbf{Aggregate.} Precomputed rollups at a coarser grain, such as daily or weekly store summaries, report marts, KPIs, budgets and forecasts, aging snapshots, and slowly changing dimension history.
\item \textbf{Metadata.} Data about data, such as lineage, dataset and model registries, identifier crosswalks, merge maps, and data-quality rules.
\end{itemize}
An LLM labels every base table from its name, columns, and row count, and a deterministic name-based classifier cross-checks the labels. The two classifiers agree on 79\% of the 1,176 base tables, with disagreements concentrated on the boundary between reference and aggregate tables (such as \texttt{int\_customer\_ltv}) or between reference and transaction tables (such as inventory exports).

Retail-1000 features a multi-layer warehouse architecture. Of its 1,004 base tables, 357 are operational source tables, 346 are vendor and acquired-system feeds (covering the acquired competitor's POS and payroll, accounting, e-commerce, delivery marketplaces, customer service, marketing, scheduling, IoT, and weather), 235 are warehouse tables across staging, intermediate, dimension, fact, aggregate, report, and snapshot layers, and 66 are frozen snapshots or sandbox copies (such as \texttt{\_old}, \texttt{\_v1}, and \texttt{\_bak} tables, analyst sandboxes, temporary tables, and external extracts). The six smaller estates consist of operational tables plus at most two frozen snapshots from schema migrations or auditor freezes.

We group documents into five types: \emph{policy and reference} documents (charters, policies, SOPs, glossaries, codebooks, handbooks, and terms), \emph{data documentation} (data dictionaries, catalogs, schema specifications, migration notes, and warehouse model guides), \emph{recurring reports} (daily, weekly, and monthly flashes, reviews, scorecards, digests, minutes, and checklists), \emph{event memos} (incident and inspection reports, launch memos, bulletins, directives, and proposals), and \emph{transcripts} of stakeholder meetings. The 1,562 documents instantiate 228 distinct templates, and recurring reports account for most dated copies.

\section{Pipeline Correctness Checks}
\label{app:pipeline_checks}

This appendix details the automated pipeline evaluator of Section~\ref{sec:eval_pipeline}. Table~\ref{tab:defect_taxonomy} defines the detection criterion and counting unit for each of the 14 error types in Table~\ref{tab:pipeline_eval}.

\begin{table}[!htbp]
\centering\small
\caption{Detection criteria and counting units for the 14 pipeline error types in Table~\ref{tab:pipeline_eval}, grouped by stage component. The LLM judge evaluates the \ul{underlined} error types, and the deterministic rule engine checks the rest.}
\label{tab:defect_taxonomy}
\setlength{\tabcolsep}{4.5pt}
\renewcommand{\arraystretch}{1.15}
\begin{tabular}{@{}>{\raggedright\arraybackslash}p{2.2cm}>{\raggedright\arraybackslash}p{3.4cm}>{\raggedright\arraybackslash}p{2.7cm}>{\raggedright\arraybackslash}p{\dimexpr\textwidth-8.3cm-6\tabcolsep\relax}@{}}
\toprule
\textbf{Component} & \textbf{Error type} & \textbf{Counting unit} & \textbf{Detection criterion and example} \\
\midrule
Compiled SP programs & \ul{Cross-column/row} \ul{dependency violation} & Dependencies & The program relaxes or violates a specified cross-table, cross-row, or intra-row dependency, e.g., sampling correlated columns independently or replacing a row formula with an aggregate. \\
 & \ul{Missing required output} & Required outputs & The program omits the operations that produce a required table, column, document, status transition, or intra-day timing window. \\
 & \ul{Missing stochastic} \ul{sampling} & Value-producing operators & The program omits a required distribution draw or replaces it with a constant, e.g., a fixed customer count where the specification requires Poisson arrivals. \\
 & Unbound variable & Variables & No execution record binds the variable, so downstream operators read an empty value. \\
 & Unintended output overwrite & Operators & The program of a recurring event type writes a fixed output name, so a later event instance overwrites a table, view, or document that the specification preserves. \\
 & Foreign key violation & Tables & After execution, rows of the table reference keys absent from the parent table, e.g., an order references a nonexistent \texttt{customer\_id}. \\
\midrule
Simulation primitives
 & Undeclared write & Operators & A write modifies a table outside the operator's declared targets. \\
 & \ul{Operator contract} \ul{violation} & Operators & The operator reports success, yet binds a value or document that contradicts its declared operands or input data, e.g., a distribution over undeclared labels or a report citing columns absent from the schema. \\
 & Placeholder value & Text columns & Materialized text values contain generator filler, e.g., \texttt{Synthetic Cafe Item\_4}. \\
\midrule
Calendar
 & Unscheduled event type & Event types & The calendar schedules no instance of the event. \\
 & Recurrence violation & Event types & Scheduled instances depart from the declared recurrence rule, collapse onto one timestamp, or violate a temporal constraint, e.g., 14 daily settlements share one minute. \\
 & \ul{Instruction mismatch} & Event types & Scheduled instances depart from the user's simulation instructions on the horizon, named dates, operating hours, or required order, e.g., restocking after trading begins. \\
\midrule
Execution
 & Database execution failure & Invocations & The invocation terminates with an unhandled database error at run time. \\
 & Unexpectedly empty table & Write targets & A table that an event type declares as a write target holds zero rows after execution. \\
\bottomrule
\end{tabular}
\end{table}

\paragraph{\textbf{Origin of the error types.}}
Each error type tests one stage guarantee of Section~\ref{sec:system}. For compiled SP programs, four error types test program validity (cross-column/row dependency violation, missing required output, missing stochastic sampling, and unbound variable), unintended output overwrites test single-program parameterization across recurring instances, and foreign key violations test referential integrity. For simulation primitives, undeclared writes and operator contract violations test that each operator stays within its declared operands below the boundary of trust, and placeholder values test that primitive execution produces no generator filler. For the calendar, unscheduled event types and recurrence violations test adherence to declared recurrence rules and temporal constraints, and instruction mismatches test adherence to the user's simulation instructions. For execution, database execution failures and unexpectedly empty tables test runtime execution and write targets. We do not score Stage~1 because users review and approve the event set during human-in-the-loop (HITL) review before compilation.

\paragraph{\textbf{Rule engine and LLM judge.}}
The evaluator inspects the scenario description, event specifications, compiled SP programs, calendar $\mathcal{T}_{\text{sched}}$, simulation log $\mathcal{L}_{\text{log}}$, database $\mathcal{D}_{\text{db}}$, and document store $\mathcal{D}_{\text{doc}}$, assigning each error type to one of two modules. A deterministic \emph{rule engine} checks the nine error types whose criteria are formal properties of program structure, calendar timestamps, execution records, or database contents. Rule checks require no LLM calls and evaluate every applicable unit. A two-pass \emph{LLM judge} (Claude Opus~5.5) evaluates the five semantic error types, whose requirements appear in natural language. For each event type, the first pass inspects the event specification, compiled SP program, calendar, and runtime evidence to propose candidate defects with verbatim evidence and specification quotes. The second pass verifies each candidate defect against four criteria: whether the cited evidence exists verbatim, whether the quoted specification requires the claimed behavior, whether the defect matches the error type's definition, and whether the defect alters analytical query answers or breaks a specified dependency.

\paragraph{\textbf{Counting units and metrics.}}
Because pipeline stages operate at different granularities, the evaluator counts each error type at the unit where the error arises. For cross-column/row dependency violations and missing required outputs, an LLM enumerates the specified dependencies and required outputs from each event specification and matches verified judge findings against them. Operator-level checks count value-producing operators (for missing stochastic sampling) or all operators in compiled SP programs (for unintended output overwrites, undeclared writes, and operator contract violations), and variable checks count declared output variables. Counting an operator defect once per compiled operator avoids inflating the rate when a recurring event executes the same operator across many calendar instances. Calendar checks count event types. Data checks count tables with foreign keys, text columns in $\mathcal{D}_{\text{db}}$, or declared write-target tables. Runtime execution checks count operator invocations because runtime failures depend on per-invocation state and loop iterations. For each error type and estate, Table~\ref{tab:pipeline_eval} reports the number of affected units over the total checked units.

\section{Data Realism Checks}
\label{app:realism}

This appendix defines the distributional metrics of Section~\ref{sec:realism} and describes the reference datasets. Table~\ref{tab:realism_full} reports the complete results, extending Table~\ref{tab:realism} with additional stylized facts and domain-specific checks. Deterministic scripts compute all metrics over the materialized SQLite databases.

\paragraph{\textbf{Reference datasets.}}
\emph{BIRD financial} is the PKDD'99 Discovery Challenge dataset of a Czech bank~\citep{berka2000guide}, redistributed in BIRD~\citep{li2023can}, containing 4,500 accounts, 5,369 clients, and 1,056,320 transactions from 1993 to 1998 with running balances. Because the Banking scenario reuses this schema, we compare the simulated and real databases column by column. \system simulated the 200 Banking clients for 120 days without accessing the real rows. \emph{BIRD debit\_card\_specializing} holds fuel-card transactions from a petrol-station chain, comprising 1,000 sampled transactions with timestamps and 383,282 monthly consumption rows across 30,545 customers. \emph{Olist}~\citep{olist2018} holds 98,202 non-cancelled orders from a Brazilian e-commerce marketplace from 2016 to 2018 with purchase timestamps, item prices, and products. \emph{Chinook}\footnote{\url{https://github.com/lerocha/chinook-database}} is the sample database whose schema the Music store scenario reuses. Its media catalog (3,503 tracks and 347 albums) contains real iTunes metadata, while a script generated its customers and invoices, so we use it as a reference for catalog metadata. We inspected each reference dataset for artifacts before selecting stylized facts. The real Czech bank books transactions on all seven weekdays, its loan payments omit interest (\texttt{payments} equals \texttt{amount}/\texttt{duration} on every loan), and its transaction amounts depart from Benford's law (against a conformity threshold of 0.015). In addition, 97\% of Olist customers place a single order, and the 1,000 fuel-card transactions span four days, so Olist does not measure repeat-customer concentration and the fuel-card sample does not measure weekly seasonality.

\paragraph{\textbf{Metrics.}}
\emph{Weekend/weekday daily volume} is the mean number of events per weekend day divided by the mean per weekday, where a value near 1 indicates a flat week. \emph{Peak 3-hour share} is the largest share of events falling within any window of three consecutive hours. \emph{Top-20\% share} sorts items by revenue or customers by activity and reports the share held by the top fifth, which the Pareto benchmark~\citep{kim2017pareto} places around 0.7--0.8. \emph{Human price endings} is the share of list prices whose cents end in .00, .25, .50, .75, .95, or .99 (including .90 for Olist, round-hundred amounts for Banking, and rates or premiums ending in .00 or .50 for Logistics and Insurance). \emph{Accounting identities} test schema-implied equations: subtotal${}+{}$tax${}={}$total and sum of lines${}={}$subtotal for receipts and invoices, net${}+{}$VAT${}={}$gross for Logistics invoices, amount due${}={}$paid${}+{}$outstanding for Insurance invoices, and, for Banking, that an end-of-day balance of each account-day equals a balance of the previous account-day plus the signed net flow of the day. \emph{Amount tail} is the ratio of the 99th to the 50th percentile of absolute amounts. \emph{Domain-specific facts} evaluate track durations, album sizes, and track prices on Music store, payment delays and VAT-rate conventions on Logistics, and claim frequency and premium concentration on Insurance. For the Banking comparison, we measure categorical mixes via total variation distance (TVD) and compare amount distribution shapes via Kolmogorov--Smirnov (KS) and Wasserstein-1 ($W_1$) distances over median-normalized $\log_{10}$ amounts.

\begin{table}[!htbp]
\centering\small
\caption{Full data-realism results extending Table~\ref{tab:realism}. Each cell reports the metric value and, in parentheses, its sample size. The top block compares Banking against BIRD \texttt{financial}, the middle block reports stylized facts across all seven estates, and the bottom block reports domain-specific facts.}
\label{tab:realism_full}
\setlength{\tabcolsep}{4pt}
\renewcommand{\arraystretch}{1.1}
\resizebox{\textwidth}{!}{%
\begin{tabular}{@{}llrrrrrrrr@{}}
\toprule
\textbf{Metric} & \textbf{Unit} & \textbf{Real reference} & \textbf{Retail-10} & \textbf{Retail-100} & \textbf{Retail-1000} & \textbf{Music store} & \textbf{Banking} & \textbf{Logistics} & \textbf{Insurance} \\
\midrule
\multicolumn{10}{@{}l}{\textbf{Banking vs.\ BIRD \texttt{financial}}}\\
Weekday mix, TVD & transactions & -- & & & & & 0.029 & & \\
Amount log-shape, KS / $W_1$ & transactions & -- & & & & & 0.236 / 0.401 & & \\
Amount median (CZK) & transactions & 2,100 & & & & & 934 & & \\
Loan amount log-shape, KS / $W_1$ & loans & -- & & & & & 0.212 / 0.122 & & \\
Loan duration / card type / statement frequency mix, TVD & rows & -- & & & & & 0.104 / 0.128 / 0.176 & & \\
Loan payment relative to amount/duration (median) & loans & 1.000 & & & & & 1.264 & & \\
Benford MAD of amounts & transactions & 0.024 & & & & & 0.012 & & \\
Final balance equals sum of transactions & accounts & -- & & & & & 180 / 180 & & \\
Balance chain holds in insertion / timestamp order & consecutive rows & -- & & & & & 0.950 / 0.475 & & \\
\addlinespace
\multicolumn{10}{@{}l}{\textbf{Stylized facts}}\\
Weekend/weekday daily volume & events & 0.75\textsuperscript{O} (98,202) 1.03\textsuperscript{B} (1,056,320) & 0.99 (2,906) & 1.02 (2,991) & 1.01 (21,553) & 1.11 (871) & 0.99 (21,500) & 1.17 (1,650) & 1.04 (3,591) \\
Peak 3-hour share & events & 0.20\textsuperscript{O} 0.24\textsuperscript{F} (1,000) & 0.42 & 0.61 & 0.79 & 0.25 & -- & -- & 1.00 \\
\quad largest single-hour share & events & 0.07\textsuperscript{O} 0.09\textsuperscript{F} & 0.17 & 0.24 & 0.74 & 0.17 & -- & -- & 1.00 \\
Top-20\% items' share of revenue & items & 0.75\textsuperscript{O} (32,729) 0.97\textsuperscript{F} (28) & 0.23 (16) & 0.25 (16) & 0.25 (16) & 0.36 (680) & -- & 0.58 (5) & 0.38 (6) \\
Top-20\% customers' share of activity & customers & 0.37\textsuperscript{B} (4,500) 0.82\textsuperscript{F} (30,545) & 0.29 (100) & 0.30 (100) & 0.34 (827) & 0.40 (94) & 0.29 (180) & 0.23 (50) & 0.37 (1,099) \\
\quad Gini of activity per customer & customers & 0.30\textsuperscript{B} 0.79\textsuperscript{F} & 0.17 & 0.19 & 0.26 & 0.35 & 0.18 & 0.06 & 0.24 \\
Share of one-line baskets / mean lines & baskets & 0.90 / 1.14\textsuperscript{O} & 0.34 / 2.10 & 0.37 / 1.99 & 0.26 / 2.63 & 0.37 / 3.23 & -- & 0.10 / 2.61 & 0.65 / 1.44 \\
Human price endings & prices & 0.85\textsuperscript{O} (112,101) 1.00\textsuperscript{C} (3,503) & 1.00 (16) & 1.00 (16) & 1.00 (16) & 1.00 (762) & 0.23 (21,500) & 0.63 (4,312) & 0.50 (6,738) \\
Accounting identities hold & rows & 1.00\textsuperscript{B} (810,268) & 1.00 (2,906) & 1.00 (2,991) & 1.00 (21,553)\textsuperscript{$\dagger$} & 1.00 (871) & 0.92 (13,915) & 1.00 (1,650) & 1.00 (7,715) \\
Amount tail, p99/p50 & amounts & 10.0\textsuperscript{O} 21.6\textsuperscript{B} 6.7\textsuperscript{F} & 3.5 & 3.2 & 2.7 & 7.5 & 15.8 & 16.9 & 6.3 \\
\addlinespace
\multicolumn{10}{@{}l}{\textbf{Domain-specific facts}}\\
Track duration p99/p50 & tracks & 10.4\textsuperscript{C} & & & & 1.7 & & & \\
Tracks per album, p90 & albums & 17\textsuperscript{C} & & & & 13 & & & \\
Track price mix (\$0.99 / \$1.99) & tracks & 0.94 / 0.06\textsuperscript{C} & & & & 0.93 / 0.07 & & & \\
Days to pay against 30-day terms (median / mean / share late) & payments & -- & & & & & & 15 / 40 / 0.41 & \\
Invoices storing the VAT rate as a fraction (0.19) / as a percentage (19.0) & invoices & -- & & & & & & 450 / 1,200 & \\
Claims per policy-year & claims & -- & & & & & & & 1.24 \\
Share of policies at the two modal premiums & policies & -- & & & & & & & 0.43 \\
\bottomrule
\end{tabular}%
}
\\[3pt]
\parbox{\textwidth}{\footnotesize \textsuperscript{B}BIRD \texttt{financial}, \textsuperscript{F}BIRD \texttt{debit\_card\_specializing}, \textsuperscript{O}Olist, \textsuperscript{C}Chinook catalog. \textsuperscript{$\dagger$}On Retail-1000, 1,018 of the 21,553 receipts belong to the six acquired stores whose POS records tax-inclusive totals (total${}={}$subtotal) as specified in the scenario, yielding 0.95 under a uniform subtotal${}+{}$tax${}={}$total rule and 1.00 once the acquired-store rule is applied.}
\end{table}

\paragraph{\textbf{Findings across domains.}}
\emph{Banking.} Beyond the transaction-code mix and intra-day timestamp ordering discussed in Section~\ref{sec:realism}, generated loans follow a 16\% annuity schedule while the 1990s reference dataset omits interest. Generated transaction amounts conform more closely to Benford's law than the real amounts, and the median-normalized log-amount distributions match within a KS distance of 0.236 on transactions and 0.212 on loans. Card type, loan duration, and statement frequency mixes match the reference within 0.104 to 0.176 TVD.
\emph{Music store.} Nearly half of the generated track titles (260 of 558 distinct titles) and 23 of 40 artists appear in the real Chinook catalog. Generated track durations have a narrower tail than the real catalog, which includes full-length television episodes and classical works, and the composer field is empty on 2\% of generated tracks compared with 28\% in Chinook.
\emph{Logistics.} The Logistics data estate records bank transactions and invoices across all seven days of the week at a constant rate (35 bank transactions per day) and bills customers at near-uniform frequencies. Customer payments arrive with realistic delays against 30-day terms (median 15 days, mean 40 days, 41\% late). Every invoice's VAT amount equals 19\% of its net total, while \texttt{vat\_rate} stores a fraction (0.19) on pre-simulation invoices and a percentage (19.0) on simulated invoices.
\emph{Insurance.} The Insurance data estate records 1.24 claims per policy-year, above typical personal-lines frequencies, and concentrates 43\% of policies on two modal premiums (\$900 and \$1,450). Loss dates spread across the day, while premium payments share an 11:00 timestamp.
\emph{Retail.} The three Retail data estates agree closely on counter-sales distributions because they share the counter-sales core. Retail-10 and Retail-100 produce a morning rush between 07:00 and 09:00 and a lunch peak, while on Retail-1000 from Day~8 onward 82--100\% of daily receipts carry a 06:00 timestamp. Basket line counts peak at two to four items because the program multiplies party size by items per person.

\section{Benchmark Quality Checks}
\label{app:benchmark_checks}

This appendix details the benchmark quality and diversity checks of Section~\ref{sec:benchmark_gen} and Section~\ref{sec:eval_complexity}. Table~\ref{tab:benchmark_quality} reports the per-estate complexity and SQL diversity metrics across all seven benchmarks.

\begin{table}[!htbp]
\centering\small
\caption{Complexity and SQL diversity of the seven benchmarks. Fractions report affected tasks over checked tasks, and Join, Grouping, and Nesting report the share of gold programs that use each SQL construct.}
\label{tab:benchmark_quality}
\setlength{\tabcolsep}{0pt}
\renewcommand{\arraystretch}{1.08}
\begin{tabular*}{\textwidth}{@{\extracolsep{\fill}}l rrrrrrr@{}}
\toprule
 & \textbf{Retail-10} & \textbf{Retail-100} & \textbf{Retail-1000} & \textbf{Music store} & \textbf{Banking} & \textbf{Logistics} & \textbf{Insurance} \\
\midrule
Doc-lookup / numeric & 5/13 & 8/10 & 6/12 & 0/10 & 1/6 & 1/6 & 4/21 \\
Requested dim.\ / complex tasks & 3/29 & 2/21 & 47/48 & 11/23 & 15/20 & 8/14 & 33/43 \\
Distinct skeletons / gold & 34/37 & 30/31 & 57/62 & 27/28 & 24/26 & 19/19 & 44/53 \\
\midrule
Join & 70\% & 48\% & 48\% & 64\% & 69\% & 58\% & 57\% \\
Grouping & 70\% & 87\% & 84\% & 64\% & 92\% & 84\% & 85\% \\
Nesting & 54\% & 55\% & 56\% & 39\% & 58\% & 74\% & 66\% \\
\bottomrule
\end{tabular*}
\end{table}

\paragraph{\textbf{Task validity checks.}}
The benchmark generator runs two deterministic validity checks on every candidate task. A task is \emph{well-formed} if its question and gold-program result are non-empty and no duplicate normalized question yields a different reference answer. A task satisfies \emph{no schema leakage} if its question names no physical table read by its gold program (exempting single-table baseline tasks, which name their target table). All 256 benchmark tasks pass both checks.

\paragraph{\textbf{Ground-truth audit.}}
The ground-truth audit checks every gold program against the four criteria of Section~\ref{sec:benchmark_gen} using $\mathcal{L}_{\text{log}}$ and $\mathcal{D}_{\text{db}}$, independently of the LLMs that derived the gold program.
\begin{itemize}[leftmargin=.5cm,itemsep=2pt]
\item \emph{Agreement with the write history.} The audit replays the write operations in $\mathcal{L}_{\text{log}}$ and counts the rows written to each table. Row-count answers match the logged write counts, and queries over a column added by a schema migration cover the rows written after that migration.
\item \emph{Authoritative tables.} The audit traces the lineage of each table read by the gold program in $\mathcal{L}_{\text{log}}$ and confirms that it reads authoritative tables, excluding deprecated tables and frozen snapshots.
\item \emph{Agreement with logged generation rules.} $\mathcal{L}_{\text{log}}$ records the column generators, value distributions, and conditional rules that populated each column. The audit confirms that these rules hold on the rows read by the gold program so that its filters and measures match the logged business logic.
\item \emph{Independence from frozen copies.} On a scratch copy of $\mathcal{D}_{\text{db}}$, the audit inserts synthetic rows into a frozen copy of each queried table and re-executes the gold program, confirming that the reference answer remains unchanged.
\end{itemize}
The benchmark generator discards any task that violates a criterion and labels each retained task as \emph{log-verified} when a criterion directly confirms the answer, \emph{corroborated} when the benchmark hint agrees with the answer, or \emph{unverified} when no criterion applies.

\paragraph{\textbf{Complexity and SQL diversity metrics.}}
Table~\ref{tab:benchmark_quality} reports three task-level ratios and three SQL structural rates for each benchmark.
\emph{Doc-lookup / numeric} counts single-measure numeric tasks whose exact reference answer also appears in a recurring report document in $\mathcal{D}_{\text{doc}}$ (such as a weekly inventory summary or monthly commission statement). Across all seven estates, 25 of 78 single-measure numeric tasks appear in a report document, while the remaining 68.0\% require querying $\mathcal{D}_{\text{db}}$.
\emph{Requested dim.\ / complex tasks} reports the share of the 198 complexity-bearing tasks whose target complexity comes from a requested disruption event, while the remainder arise from status codes, discount policies, and dated snapshots introduced by routine operational events.
\emph{Distinct skeletons / gold} counts unique normalized SQL syntax trees after replacing table, column, and literal tokens with placeholders, confirming that composing verified query fragments produces structurally diverse gold queries (235 distinct skeletons across 256 tasks, or 91.8\% overall) that combine multi-table joins (59.3\%), grouping and aggregation (81.0\%), and nested subqueries or CTEs (57.4\%).

\end{document}